%% file: main.tex
\documentclass[conference]{IEEEtran}
\IEEEoverridecommandlockouts

\usepackage{cite}
\usepackage{amsmath,amssymb,amsfonts}
\usepackage{algorithm}
\usepackage{algpseudocode}
\usepackage{graphicx}
\usepackage{textcomp}
\usepackage{booktabs}
\usepackage{multirow}
\usepackage{adjustbox}
\usepackage[table]{xcolor}
\usepackage{colortbl}
\def\BibTeX{{\rm B\kern-.05em{\sc i\kern-.025em b}\kern-.08em
    T\kern-.1667em\lower.7ex\hbox{E}\kern-.125emX}}

\usepackage{colortbl}
\definecolor{lightblue}{RGB}{173, 216, 230}
\definecolor{lightgreen}{RGB}{144, 238, 144}
\definecolor{lightgrey}{RGB}{211, 211, 211}

\DeclareMathVersion{tabularbold}
\SetSymbolFont{operators}{tabularbold}{OT1}{cmr}{b}{n}

\usepackage{pgfplotstable} 
\usepackage{algpseudocode}

\graphicspath{ {./images/} }

\newtheorem{definition}{Definition}

\definecolor{lightblue}{RGB}{173, 216, 230}
\definecolor{lightgreen}{RGB}{144, 238, 144}
\definecolor{lightgrey}{RGB}{211, 211, 211}

\newcommand{\ens}{\enspace}

\pgfplotstableset{
    /color cells/min/.initial=0,
    /color cells/max/.initial=1000,
    /color cells/textcolor/.initial=,
    color cells/.code={%
        \pgfqkeys{/color cells}{#1}%
        \pgfkeysalso{%
            postproc cell content/.code={%
                \begingroup
                \pgfkeysgetvalue{/pgfplots/table/@preprocessed cell content}\value
                \ifx\value\empty
                    \endgroup
                \else
                    \pgfmathfloatparsenumber{\value}%
                    \pgfmathfloattofixed{\pgfmathresult}%
                    \let\value=\pgfmathresult
                    \pgfplotscolormapaccess
                    [\pgfkeysvalueof{/color cells/min}:\pgfkeysvalueof{/color cells/max}]%
                    {\value}%
                    {\pgfkeysvalueof{/pgfplots/colormap name}}%
                    
                    \pgfkeysgetvalue{/pgfplots/table/@cell content}\typesetvalue
                    \pgfkeysgetvalue{/color cells/textcolor}\textcolorvalue

                    \edef\escapedvalue{\noexpand\detokenize\expandafter{\typesetvalue}}

                    \toks0=\expandafter{\escapedvalue}%
                    \xdef\temp{%
                        \noexpand\pgfkeysalso{%
                            @cell content={%
                                \noexpand\cellcolor[rgb]{\pgfmathresult}%
                                \noexpand\definecolor{mapped color}{rgb}{\pgfmathresult}%
                                \ifx\textcolorvalue\empty
                                \else
                                    \noexpand\color{\textcolorvalue}%
                                \fi
                                \the\toks0
                            }%
                        }%
                    }%
                    \endgroup
                    \temp
                \fi
            }%
        }%
    }
}
\pgfplotsset{compat=1.18}

\usepackage{caption}

\begin{document}

\title{Tailored Architectures for Time Series Forecasting: Evaluating Deep Learning Models on Gaussian Process-Generated Data\\
\thanks{This work has been accepted at IJCNN 2025.}
}

\author{\IEEEauthorblockN{Victoria Hankemeier}
\IEEEauthorblockA{\textit{Autonomous Intelligent Systems Group} \\
\textit{University of Münster}\\
Münster, Germany \\
v.hankemeier@uni-muenster.de}
\and
\IEEEauthorblockN{Malte Schilling}
\IEEEauthorblockA{\textit{Autonomous Intelligent Systems Group} \\
\textit{University of Münster}\\
Münster, Germany \\
malte.schilling@uni-muenster.de}
}

\maketitle

\begin{abstract}
Developments in Deep Learning have significantly improved time series forecasting by enabling more accurate modeling of complex temporal dependencies inherent in sequential data. The effectiveness of such models is often demonstrated on limited sets of specific real-world data. Although this allows for comparative analysis, it still does not demonstrate how specific data characteristics align with the architectural strengths of individual models. Our research aims at uncovering clear connections between time series characteristics and particular models. We introduce a novel dataset generated using Gaussian Processes, specifically designed to display distinct, known characteristics for targeted evaluations of model adaptability to them. Furthermore, we present \textit{TimeFlex}, a new model that incorporates a modular architecture tailored to handle diverse temporal dynamics, including trends and periodic patterns. This model is compared to current state-of-the-art models, offering a deeper understanding of how models perform under varied time series conditions.
\end{abstract}


\section{Introduction}
Real-world data is often sequential, comprising ordered series of data points collected over time. Such time series data, while originating from diverse domains, typically exhibits several common characteristics. A prominent example are reoccuring patterns such as higher traffic activities at the start and end times of a working day. Time series forecasting is a fundamental task in various domains which aims at predicting future values based on past observations. However,  uncovering temporal patterns is challenging due to the multitude of interacting patterns and influences that act on different time scales, regardless of the domain. Advanced modeling techniques are required to address non-stationarity, i.e. shifts in mean, variance, and auto-correlation over time. Moreover, incorporating long-term dependencies, which reveal the impact of distant past events on present and future outcomes, can be computationally expensive. Hence, balancing accuracy and computational complexity as a crucial task of the development of time series forecasting models.

Recent advancements have led to the creation of various domain-independent deep learning models for long and short-term time series forecasting. However, these models have predominantly been evaluated  using
 the same limited number of real-world benchmark datasets. While this approach demonstrates the models’ effectiveness on real-world data and facilitates comparison in specific application areas, it also leads to a wide spectrum of models performing at comparable levels on many of these benchmark tasks. What is missing is an explanation on differences between models in the respective benchmarks: Is there a reason one model is performing well or better in one, but not another benchmark? There is currently no comprehensive analysis on how these models' specific design choices relate to particular characteristics of time series data. To close this research gap we introduce an artificial dataset created using various kernel functions as covariance functions of Gaussian Processes (GPs). The time series are then sampled from these GPs, ensuring that they exhibit specific dominant characteristics as specified by the kernel functions. Training models on this dataset allows for a comprehensive evaluation of the models and their architecture, enabling not only performance comparisons against general datasets but also detailed analyses of their specific advantages and limitations when applied to data with distinct characteristics.
 We further develop a deep learning model with a modular architecture to systematically evaluate commonly used components through an ablation study. This study uses the sampled dataset as well as widespread benchmarking datasets. To ensure comparability, we maintain a well-structured and reproducible training and evaluation codebase, aligned with previously established libraries and publications.
 Our main contribution is two-fold:
\begin{itemize}
    \item  We provide a set of sampled datasets with guaranteed data characteristics using Gaussian Processes. Performing a nuanced analysis of deep learning models trained on those datasets we extensively evaluate their architectural design choices.
    \item With \textit{TimeFlex} we developed a flexible yet efficient model that processes the trend in the time domain to capture changes over time in the distribution and applies Fourier Transform on the seasonal component to better capture periodic patterns. Convolution Layers with exponentially increasing dilation rates capture periodic patterns at different lengths scales. The modular structure enables the methodical assessment of individual components.
\end{itemize}

\section{State-of-the-Art and Related work}
 There is a wide spectrum of methods used for time series forecasting. We will focus on deep learning based architectures and differentiate these, at first, with respect to their basic building blocks. The different types of neural network approaches range from simple Multi-Layer Perceptrons (MLPs) to Transformer architectures and, further, include models with an explicit internal state as in Recurrent Neural Networks (RNNs). These approaches differ considerably in their architectural design and computational complexity. Still, many of current State of the Art (SotA) time series forecasting models share some commonalities: Foremost many explicitly distinguish—and try to model these individually—a periodic recurring temporal pattern from the general non-stationary trend. We will present an overview of SotA models that can be categorized in MLPs, RNNs, Transformers, Convolution Neural Networks (CNNs), State Space Models (SSMs), and Mixed Architecture Models. A summary is given in Table \ref{tab: SotA}.

\subsection{SotA in Deep Learning for Time Series Forecasting}

\paragraph{MLP} The most elementary structures are MLPs, which consist of an input layer, one or more hidden layers and an output layer, processing a fixed-length sequence of time steps. Despite their relatively fundamental structure, they demonstrate a high level of competitiveness, especially when decomposing the input into a trend and seasonal part as in \textit{DLinear} \cite{DLinear}. Variants of this approach consider normalization such as subtracting and then re-adding the last value of the sequence as in \textit{NLinear} \cite{DLinear}, or reverse instance normalization (RevIN) with a linear layer, as in the \textit{RLinear} model\cite{RLinear}. All of these models achieve high performance with low computational demand. The architecture of \textit{DLinear} was enhanced in \textit{WaveMask/WaveMix} \cite{waveaug2024} by applying data augmentation methods before the \textit{DLinear} architecture using five different augmentation types with the aim of more effectively capturing periodicity. \textit{TimeMixer} \cite{TimeMixer} decomposes the series at multiple scales and uses a Past-Decomposable-Mixing (PDM) block for bottom-up mixing of past seasonal information and top-down for past trend information and a Future-Multipredictor-Mixing (FMM) block for aggregation. In \textit{TimesNet} \cite{TimesNet} each period's intra-period and inter-period variations are captured simultaneously by a block transforming the 1D series into a 2D space.

\paragraph{RNN} Since MLPs do not innately process temporal sequences, RNNs were developed to enhance learning of sequential relationships by using a hidden state that integrates the current input with the hidden state from the previous time step. However, RNNs face challenges such as the vanishing gradient problem during back-propagation over long sequences which impedes learning long-term dependencies. To overcome this issue, gating mechanisms as in Long Short-Term Memory (\textit{LSTM}) \cite{LSTM} and Gated Recurrent Units (\textit{GRU}) \cite{GRU} units were proposed. These gates manage the flow of information, deciding what part of the internal state to retain, update, or forget, and have become the standard for years for time series forecasting due to their effectiveness.

\paragraph{CNN}  Although RNNs effectively capture temporal dynamics, they can be slow to train, particularly when dealing with long time series. The introduction of CNNs into time series analysis was a significant advancement in handling time-dependent data more efficiently. The first to apply CNNs on time series were Zheng et al. \cite{CNN} who proposed the Multi-Channels Deep Convolution Neural Networks (\textit{MC-DCNN}) for classification. Building on this concept, Oord et al. \cite{WaveNet} developed \textit{WaveNet}, employing dilated causal convolutions to extract periodic patterns across different scales, which was originally developed for generating raw audio waveforms but also proved effective in time series forecasting. Lea et al. \cite{TCN} adapted the \textit{WaveNet} architecture for time series forecasting, i.e. the Temporal Convolutional Network (\textit{TCN}) with a hierarchy of temporal convolutions. In \textit{AdaWaveNet} \cite{Adawavenet} a decomposition layer splits trend and seasonal components. The seasonal component is then transformed using the Lifting Wavelet Transform (LWT) at different scales instead of dilated convolutions as used in other models.

\paragraph{Transformer} The concept of attention has driven the success of natural language processing. It has been adapted for time series analysis, particularly using Transformer architectures. When initially adopting Transformers on time series, Li et al. \cite{Li_transformer} identified main bottlenecks: the lack of position sensitivity in dot-product self-attention and the quadratically growing complexity over sequence length. In \cite{Informer} the ProbSparse self-attention mechanism was introduced and a generative-style decoder enabled long sequence output in a single forward step, both to reduce complexity. The \textit{Autoformer} \cite{Autoformer} was the first architecture to incorporate a decomposition layer and replace traditional self-attention calculation by performing auto-correlation on the Fast Fourier Transform (FFT). \textit{FEDformer} \cite{FEDFormer} improved this by using Fourier and Wavelet enhanced blocks to highlight the most significant frequency components, employing the convolution theorem for more efficient representation learning by using point-wise multiplication in the frequency domain instead of time-domain convolutions. In \textit{Pyraformer} \cite{Pyraformer} a Pyramidal Attention Module with a pyramidal graph structure models the temporal dependencies at multiple resolutions for enhanced efficiency when handling short- and long-term dependencies. \textit{Crossformer} \cite{Crossformer} explicitly exploits cross-dimension dependencies for multi-variate time series and \textit{PatchTST} \cite{PatchTST} processes subseries-level patches rather than point-wise input. 
Instead of encoding multivariate data at each time step, in \textit{iTransformer} \cite{Itransformer} independent time series are used as variate tokens. Yang et al. \cite{VCFormer} proposed \textit{VCFormer}, using inverted encoding from \cite{Itransformer}, enhancing \cite{Autoformer}'s auto-correlation with Variable Correlation Attention, and applying FFT-based query-key multiplication. The Koopman Temporal Detector handles non-stationarities via linear projections. In \textit{TimeXer} \cite{wang2024timexer}, endogenous variables are embedded with a learnable global token, followed by temporal self-attention, cross-attention with exogenous variate tokens, normalization, and feedforward layers.

\paragraph{SSM}\textit{Mamba} \cite{Mamba} is a recent SSM restructuring common SSM architectures combining it with a gated MLP block that demonstrated state of the art results. In \textit{TimeMachine} \cite{TimeMachine}, two pairs of Mamba blocks (forward and transposed) efficiently capture long-term dependencies, enabling both independent and mixed channel processing. Similarly, \textit{SMamba} and \textit{DTMamba} use Mamba block pairs. \textit{SMamba} captures mutual information among variates, transforming inputs with a linear tokenization layer and learning temporal dependencies via a Feed Forward Temporal Dependencies Encoding Layer. \textit{DTMamba} applies RevIN and a channel independence layer, with its Twin-Mamba block capturing low- and high-level temporal patterns.

\paragraph{Models using Prior Knowledge}
The Time-series Dense Encoder (TiDE) in \cite{TiDE} utilizes covariates by encoding the time series along with these covariates, then decoding the time series with future projections of the covariates. Both the encoder and decoder consist of residual blocks that include activation functions, a dense linear layer, and a dropout function for regularization. \textit{SparseTSF} \cite{Sparsetsf} samples the original series down into $w$ sub-sequences, where $w$ is the known periodicity and combines it with a linear layer. In \textit{CycleNet} \cite{Cyclenet}, learnable recurrent cycles are used to model the inherent periodic patterns within sequences, which are combined through a single-layer linear and a dual-layer MLP backbone to enhance learning efficiency and prediction accuracy.

\paragraph{Mixed-Model Architectures} 
To combine the strengths of different models, hybrid approaches like Mixture of Universals (\textit{MoU}) \cite{MoU} integrate adaptive feature extraction with parts of multiple architectures. In their approach, learnable router patches direct inputs to be sequentially processed through a Mamba layer, feedforward layers, and convolution layers to enhance non-linearities and receptive fields. Long-term dependencies are captured via softmax self-attention. Similarly, \textit{OneNet} \cite{Onenet} employs a two-stream architecture (cross-time and cross-variable), merging outputs with online convex programming and learnable ensemble weights.

\subsection{Related Work on Surveys and Benchmarking Frameworks}
As there is a multitude of various domain-independent deep learning models for both long and short-term forecasting, a critical review of these models' performance across standard benchmarks is provided in \cite{wang2024survey}. They present a comparison of SotA models on different time series tasks and data sets and a widely used library for benchmarking purposes.The evaluations often fail to reveal a clear connection between a model's architectural nuances and specific characteristics of the datasets, which limits the ability to explain why a certain model performs better in specific scenarios. Qiu et al. \cite{qiu2024tfb} identified this absence of a fair and unbiased comparison, which motivated them to evaluate some of the models on more datasets from a larger variety of domains. They skipped the drop-last method, grouped datasets by key features, and used Seasonal-Trend Decomposition using Loess (STL), Augmented Dickey-Fuller test (ADF), Pearson correlation coefficients (PCCs), and transition algorithms for analysis. Models were compared across groups, but no qualitative analysis linked architectures to features, and measures were highly parameter-dependent.

\begin{table*}[t]
\centering
\resizebox{\textwidth}{!}{
\begin{tabular}{|c|c|c|c|c|c|c|c|c|}
\toprule
\textbf{Category} & \textbf{Model} & \textbf{Characteristics} & \textbf{CI/CD} & \textbf{Prior Knowledge} & \textbf{Decomposition} & \textbf{Transform} & \textbf{Norm} & \textbf{Complexity}\\
\midrule
\multirow{9}{*}{\textbf{MLP}} & DLinear & Moving avg., FC layer & CI or CD & - & T+S & - & - & simple \\
\cline{2-9}
& RLinear & Linear mapping & CI or CD & - & - & - & RevIN & simple \\
\cline{2-9}
& NLinear & Linear mapping & CI or CD & - & - & - & BN & simple \\
\cline{2-9}
& TiDE & MLP-based encoder-decoder & CD & future covariates & - & - & - & simple \\
\cline{2-9}
& TimeMixer & MLP-based, multi-scale mixing & Both & - & Multi-scale T+S & - & z-Score or subtract last & simple\\
\cline{2-9}
& Wave-Mask/Mix & Wavelet-Mixing \& DLinear & CI or CD & - & T+S & several & - & simple\\
\cline{2-9}
& SparseTSF & Sparse forecasting with MLP or linear layer & CD & Period length & - & - & SN & very simple \\
\cline{2-9}
& CycleNet & learnable recurrent cycles & CI & Cycle length & Cycle subtraction & - & RevIN, opt. & simple \\
\midrule
\multirow{3}{*}{\textbf{Convolution}} & TimeFlex & Linear mapping, dilated conv. & CI or CD & - & T+S & FFT, opt. & RevIN, opt.  & simple \\
\cline{2-9}
& WaveNet & Dilated Convolutions & CI or CD & - & - & - & - & simple \\
\cline{2-9}
& AdaWaveNet & Clustering of trend, 1D conv. & CI & - & T+S & Wavelet (lifting scheme) & RevIn, opt. & mid \\
\midrule
\multirow{7}{*}{\textbf{Transformers}} & VCFormer & Cross-correlation attention & CI & - & - & FFT for attention  & z-Score, opt. & very high \\
\cline{2-9}
& PatchTST & Patched input, linear projection & CI & - & T+R  & - & RevIN, opt.& high \\
\cline{2-9}
& iTransformer & Inverted token & CI & - & -  & - & z-Score, opt. & high \\
\cline{2-9}
& Informer & Vanilla time series transformer & CD & - & - & - & - & high \\
\cline{2-9}
& Autoformer & Auto-correlation on FFT & CD & - & T+S  & FFT for attention & - & very high \\
\cline{2-9}
& FEDformer & MOE decomp., like Autoformer & CD & - & T+S & FFT/DWT & - & very high \\
\cline{2-9}
& Crossformer & Cross-time and cross-dim. attention & CI & - & Two-stage attention & - & LN & very high \\
\midrule
\multirow{3}{*}{\textbf{State Space}} & SMamba & Bi-directional Mamba blocks & CD & - & - & - & LN & very high \\
\cline{2-9}
& DTMamba & 2 TwinMamba blocks & CI & - & Twin  & - & RevIN & very high \\
\cline{2-9}
& TimeMachine & Quadruple Mambas & Both & - & - & - & RevIN  & very high \\
\midrule
\multirow{2}{*}{\textbf{Mixture}} & MoU & Sequential application of architectures & CI & - & - & - & RevIN  & very high \\
\cline{2-9}
& OneNet & Online Ensemble cross-time \& -feature & CI & - & optional & optional & optional & very high \\
\bottomrule
\end{tabular}
}
\caption{Summary of SotA models comparing key characteristics.}
\label{tab: SotA}
\end{table*}

\section{Fundamentals of time series data} \label{fundamentals}

We define a stochastic process for a probability space $(\Omega, \mathcal{F}, \mathbb{P})$, where $\Omega$ is the sample space, $\mathcal{F}$ the event space and $\mathbb{P}$ the probability measure, as follows:

\begin{definition} [Stochastic Process]
    A stochastic process $X$ with index set $T$ is a collection of random variables $X=\{ X_t : t \in T \}$ defined on $(\Omega, \mathcal{F}, \mathbb{P})$. The function $t \mapsto X_t (\omega)$, with $\omega \in \Omega$, is called the realization of $X$.
\end{definition}
Following standard practice, we employ weak stationarity.
\begin{definition}[Weak Stationarity]
    A stochastic process $X$ is said to be weakly stationary if, \\
    $\mathbb{E}[X_t] = \mu$, $\mathrm{Var}(X_t) = \sigma^2 < \infty$, $\mathrm{Cov}(X_t, X_{t+h}) = \gamma(h), \\
    \quad \text{for all time lags} \enspace h \in \mathbb{Z}$ hold for all $t \in T$.
\end{definition}

\begin{definition} [Time series]
    A time series is a specific realization of a stochastic process with an index set $T$ defining the range of time $t$.
\end{definition}

When performing time series forecasting we only consider realizations of the filtration $\mathbb{F}:=\{F_t\}_{t \geq 0}$, which takes the observations from a starting time $t=0$ up to time $t$ and we assume observations to be equally spaced in time. 

As realizations of stochastic processes, time series are strongly characterized by their underlying dynamics. Stationary time series exhibit recurring patterns with defined periodicity that remain consistent over time. Such pattern exhibit short-time auto-correlation which makes them easy to learn. Real-world time series typically exhibit long- and short-term trends, distributional shifts, and random fluctuations, making them inherently non-stationary across different temporal scales. Besides, for a measured stochastic process $Y_t$ there is a measurement noise $\epsilon_t$ present as $Y_t = X_T + \epsilon_t$. State-of-the-art (SotA) methods show differing strengths and weaknesses. Methods that are effective for addressing some characteristics might face limitations when applied to time series with differing attributes.

\section{Dataset Generation using Gaussian Processes}

Recent methods for time series prediction aim to handle varying temporal scales, trends, and non-stationarity. As previously described, proposed measures for dataset characteristics highly depend on the parameter selection highlighting the need for alternative evaluation methods that are less sensitive to such variations. Therefore, we present as a novel evaluation for models a novel dataset generated from Gaussian Processes. Using GPs allows us to control the characteristics of the generated data and key underlying statistical patterns, thereby providing a robust framework for assessing the strengths and limitations of proposed time series forecasting models.

\subsection{Gaussian Processes}

\begin{definition}\label{def:GP}A Gaussian Process is a stochastic process, such that the joint distribution of any finite subset of random variables of this process is distributed according to a multivariate Gaussian distribution \cite{RasmussenWilliams2005}.
\end{definition}

A GP can be described as a distribution over functions, where the random variables represent the function values $f(x)$ at input locations $x$ from a set $\{x_i\}_{i=1}^N, \quad \text{where } x_i \in T \text{ for } i = 1, \ldots, N$,  representing $N$ samples. The function $f(x)$ is distributed according to:

\begin{equation}
	f(x)\ens \sim \ens \mathcal{GP}(m(x),k(x_p,x_q)), \ens \text{for} \ens x_p, x_q \in \{x_i\}_{i=1}^N
\end{equation}
where
\begin{equation}
	\begin{split}
	m(x) &= \mathbb{E}[f(x)],\\
	\text{and}\quad k(x_p, x_q) &= \mathbb{E}[(f(x_p)-m(x_p))(f(x_q)-m(x_q))].
	\end{split}
\end{equation}
Here, $\mathbb{E}$ describes the expected value. Without loss of generality, we assume a zero mean in the following. GPs offer a high flexibility as different covariance functions can be selected, making them well suited for our dataset creation. 

\subsection{Dataset Generation} \label{sec:data_gen}
We generated samples from GPs for an input vector $\textbf{x}_N = [ x_1, \ldots, x_N ]$ for $N$ samples and $x_i \in T$ with a covariance matrix:
\begin{equation}
    \text{Cov}(x_p, x_q)=k(x_q, x_p)+ \epsilon I, \quad \text{for } p, q = 1,\ldots, N
\end{equation} 
using kernel functions $k$ and an additional noise term $\epsilon$ representing measurement noise. As kernel function, we use the stationary Gaussian (SE), periodic and rational quadratic kernel, as well as non-stationary locally periodic, and combined kernel functions:
\begin{align}
k_{\text{se}}(x_p, x_q) &= \exp\left(-\frac{(x_p - x_q)^2}{2l^2}\right), \\
k_{\text{periodic}}(x_p, x_q) &= \exp\left(-\frac{2\sin^2\left(\frac{\pi |x_p - x_q|}{\tau}\right)}{l^2}\right)\\
k_{\text{locally periodic}}(x_p, x_q) &= 
k_{\text{se}}(x_p, x_q) * k_{\text{periodic}}(x_p, x_q),\\
k_{\text{rational quadratic}}(x_p, x_q) &= 
\left(1 + \frac{(x_p - x_q)^2}{2\alpha l^2}\right)^{-\alpha},\\
k_{\text{combined}}(x_p, x_q) &= 
k_{\text{se}}(x_p, x_q) + k_{\text{periodic}}(x_p, x_q),
\end{align}
with $l$ representing the length scale, $\tau$ the period, and $\alpha$ the shape parameter. The kernels are configured with $l=0.5$, $\tau = 24$, $\alpha = 1.0$, and $\epsilon = 1 \times 10^{-10}$.
\\
Applying the SE-kernel results in data with a fixed average distance from the mean, and its slowly varying fluctuations make it well-suited for trend modeling. The periodic kernel results in a periodic pattern, as commonly observed in real-world data with inherent seasonality. The locally periodic kernel is a multiplication of the SE-kernel and the periodic kernel, and reflects periodic patterns that change over time, modeling a trend or a smooth distribution shift. The rational quadratic kernel is a composition of several SE-kernels with different length scales, resulting in fluctuations at different time scales. The dataset sampled from the combined kernel merges short periodic patterns with long-term recurring patterns. We sampled data points with an hourly timestamp over a span of one year. For each kernel, four samples were created to produce a multi-variate time series with four distinct features. By sampling from the same kernel with consistent parameters, we ensure correlation among the features. Using this variety of kernel functions enables us to capture typical real-world characteristics that may not be explicitly separable or uniquely identifiable in observed data, but are nonetheless present as overlaid patterns. The dataset \textit{GP-TimeSet} along with the corresponding code is published for further use and comparison of models\footnote{see \textit{https://github.com/vicky-hnk/time-flex}}.

\section{TimeFlex Model Architecture}
We also propose a flexible and modular architecture: The \textit{TimeFlex} model integrates several of the proposed approaches and is designed to address the various identified characteristics of temporal data. As such, frequency-domain transformations and dilated convolutions maximize the model's ability to learn seasonal dynamics, while linear dense layers ensure robust trend representation. 

\subsection{Overview Architecture}
\begin{figure}
    \centering
    \includegraphics[width=1\columnwidth]{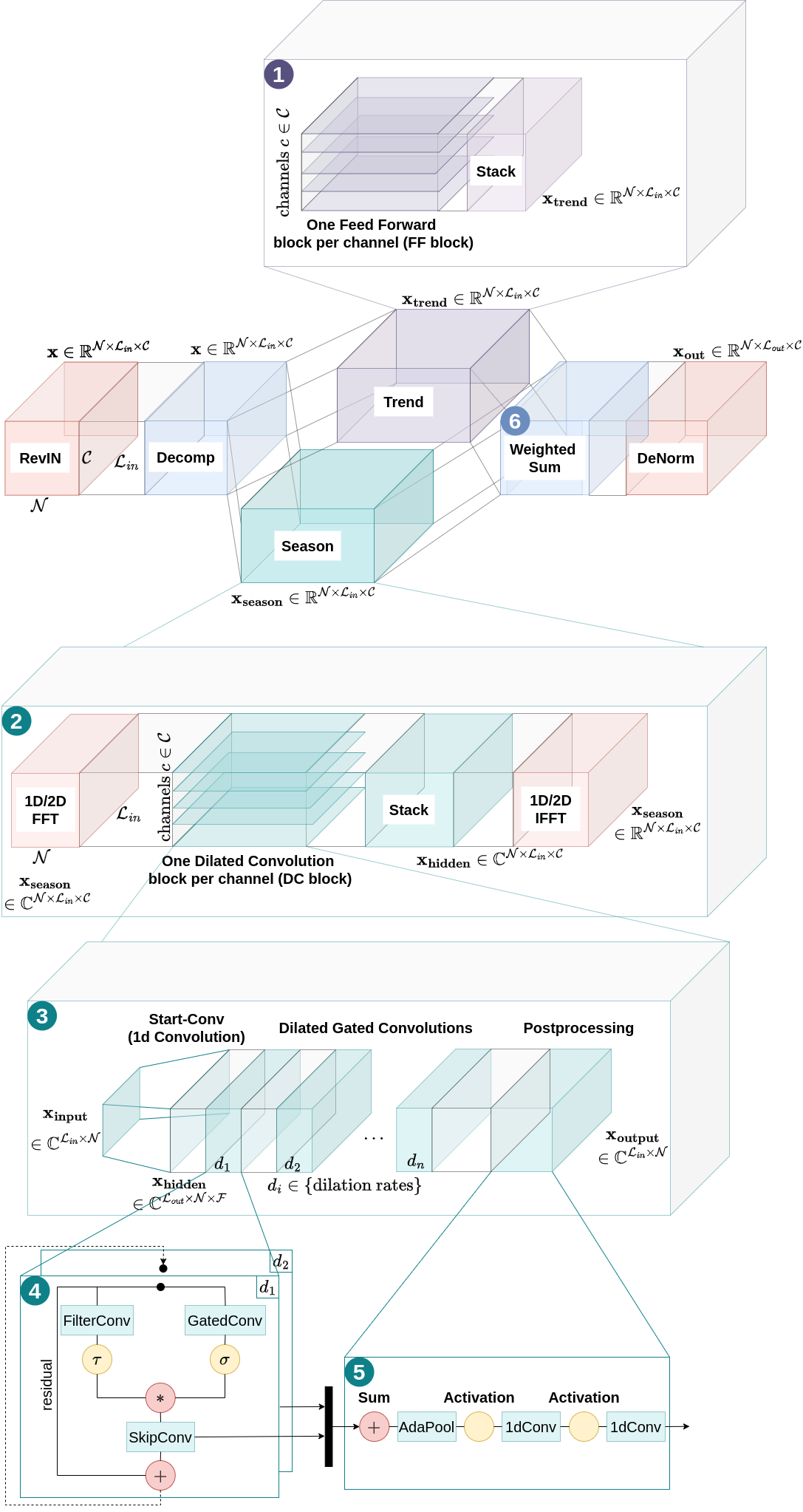} 
    \caption{Architecture of TimeFlex model. The red boxes are optional parts.}
    \label{fig:overall}
\end{figure}

The overall architecture to process one batch is illustrated in Fig. \ref{fig:overall}. For simplicity only the model variant with channel-independent (CI) processing is depicted, a channel-dependent (CD) processing is equally viable. The red boxes in the figure are optional.

\paragraph{Normalization and Decomposition}
The input tensor is $\mathbf{x} \in \mathbb{R}^{\mathcal{N} \times \mathcal{L}_{in} \times \mathcal{C}}$ with $\mathcal{N}$ being the number of sequences, $\mathcal{L}_{in}$ the input sequence length and $\mathcal{C}$ the number of variates/channels (shown in left, top part of Fig. \ref{fig:overall}). As a first optional step, the input may be normalized using RevIN \cite{kim2022reversible}. For a time step $t$, a variate $c\in\mathcal{C}$ and an instance, i.e. a sequence, $i\in \{1,..., \mathcal{N}\}$ this normalization would be computed as:
\begin{equation}
x^{(i)}_{norm,ct} = \gamma_c \left( \frac{x^{(i)}_{ct} - \mathbb{E}_t[x^{(i)}_{ct}]}{\sqrt{\text{Var}[x^{(i)}_{ct}] + \epsilon}} \right) + \beta_c,
\end{equation}
where $\gamma, \beta \in \mathbb{R}^C$ are learnable affine parameter vectors and the mean and variance are calculated as:

\begin{eqnarray}
\begin{split}
\mathbb{E}_t[x^{(i)}_{ct}] &= \frac{1}{\mathcal{L}} \sum_{j=1}^{\mathcal{L}} x^{(i)}_{cj}, \\
\quad \text{Var}[x^{(i)}_{ct}] &= \frac{1}{\mathcal{L}} \sum_{j=1}^{\mathcal{L}} \left(x^{(i)}_{cj} - \mathbb{E}_t[x^{(i)}_{ct}]\right)^2.
\end{split}
\end{eqnarray}

The trend and seasonal components are extracted via moving average decomposition \cite{Autoformer}. The input tensor is padded to allow a sliding window of size $w$ to compute the trend $T_t$ and seasonal component $S_t$:
    \begin{equation} T_t = \frac{1}{w} \sum_{i=t- [ w/2 ]}^{t + [ w/2 ]} x_{\text{padded}, i}, \ens S_t = x_t - T_t 
    \end{equation}
\paragraph{Trend Processing}
The trend component is processed by a feedforward layer (see Fig. \ref{fig:overall} -- box 1).   
When channels are modeled independently, separate linear layers are employed for each channel, as detailed in Equation \ref{eq:trend_combined}, first part ($\mathbf{x_{trend, c}}$). In order to capture channel dependencies, one has to apply one linear layer, as described in Equation \ref{eq:trend_combined}, second part ($\mathbf{x_{trend}}$).

\begin{align}
    \mathbf{x_{trend, c}} = \mathbf{W}_c \mathbf{x}_c + \mathbf{b}_c , \quad \mathbf{x_{trend}} = \mathbf{W} \mathbf{x} + \mathbf{b}. \label{eq:trend_combined}
\end{align}

\paragraph{Season Processing}

The structure of the seasonal part is shown in box 2 in Fig. \ref{fig:overall}. To effectively capture periodic or locally periodic patterns, the seasonal part is transformed by a Discrete Fourier Transform (DFT) using the Fast Fourier Transform. The DFT $X_k$ of $x_{season}$ is given by:
\begin{equation}
    X_{k,c}^{(i)} = \sum_{t=0}^{\mathcal{L}-1} x_{\text{season},ct}^{(i)} e^{-i \frac{2\pi}{\mathcal{L}} k t}
\end{equation}
Note, that we ran our experiments using 1D-FFT across the time dimension as well as 2D-FFT across the time and variate dimension, enabling information exchange between channels.

Analogously to the trend component, the model employs either a separate Dilated Convolution (DC) block for each variate or a single DC block processing all variates jointly, facilitating enhanced information exchange between variates. A DC block for one channel for the CI variant is depicted in Figure \ref{fig:overall} -- box 3. A similar DC block is applied within the CD variant, processing a tensor with a third (channel) dimension. The DC blocks are structured similarly to \cite{WaveNet} and apply the Gated Convolutions from \cite{PixelCNN}. The DC blocks are formally defined in the following equations \ref{eq: DC} to \ref{eq:dc last}.

\begin{equation} \label{eq: DC}
\mathbf{x} = \left[ \text{DC}_1(\mathbf{x}_1), \dots, \text{DC}_c(\mathbf{x}_c) \right].
\end{equation}
In every DC block a 1-D convolution layer is applied to transform the input $x_{season}$ to a higher feature space:
\begin{equation}
      \mathbf{x}_{hidden}= \text{CausalConv}(\mathbf{x}_{\text{season}})
\end{equation}

For every dilation rate $d$ in a set of dilation rates $\{1, ..., \mathcal{D}\}$ the following Gated Convolutions are applied (see Figure \ref{fig:overall} -- box 4):

\begin{align}
    \quad \mathbf{h}_{\text{gau},d} &= \tanh \left( W_{f,d} \ast \mathbf{x}_{hidden} \right) \odot \sigma \left( W_{g,d} \ast \mathbf{x}_{hidden} \right) \\
    \quad \mathbf{h}_{\text{skip},d} &= \text{Conv} \left( \mathbf{h}_{\text{gau},d} \right) \\
    \quad \mathbf{x}_{d+1} &= \mathbf{x}_d + \mathbf{h}_{\text{gau},d} \\
    \quad \mathbf{x}_{skip} &= \mathbf{x}_{skip} + \mathbf{h}_{\text{skip},d} \label{eq:dc last}
\end{align}

where $*$ denotes a convolution operator, $\odot$ an element-wise multiplication operator, $\sigma(\cdot)$ a sigmoid function, $f$ and $g$ filter and gate, respectively, and $W$ a learnable convolution filter. The dilation rates have exponential growth to widen the receptive field and better capture long-term dependencies. After stacking the seasonal components to one tensor, further post-processing including Adaptive Pooling (Pool), activation functions (Act), dropout layers (Drop) and Convolutions (Conv) is applied to capture additional non-linearities and adjust the output dimension of $x_{s,out}$:

\begin{equation} \label{eq: postproc}
\mathbf{x}_{s, out} = \text{Conv}(\text{Drop}(\text{Act}(\text{Conv}(\text{Act}(\text{Pool}(\mathbf{x}_{skip})))))).
\end{equation}

This is represented in Figure \ref{fig:overall} -- box 5. In the frequency domain, the tensor $x_{season}$ is complex-valued ($x_{season} \in \mathbb{C}^{\mathcal{N}\times \mathcal{L} \times \mathcal{C}}$) and the layers are thus implemented as complex layers and activation functions using complex multiplication:
\begin{equation}
    \text{Layer}(x) = \big(L_r(x_r) - L_i(x_i)\big) + i \big(L_r(x_i) + L_i(x_r)\big)
\end{equation}
with $L_r$ and $L_i$ being the real and imaginary layers and $x_r$ as well as $x_i$ the real and imaginary parts of $x_{season}$, respectively. For the seasonal part within the frequency domain, the inverse DFT is applied to convert the output back to the time domain. 

\paragraph{Weighted Sum}
Finally, trend and seasonal components are recombined using a weighted sum (shown in the top right part of Fig. \ref{fig:overall}, marked as step 6).
\begin{equation}
x_{out} = w_{season} \cdot x_{season, out} + w_{trend} \cdot x_{trend, out}
\end{equation}
$w_{\text{season}}$ and $w_{\text{trend}}$ are learnable weights initialized to $1$ and updated during training to control the contribution of the seasonal and trend components, respectively.
During our ablation studies, we also performed experiments applying the post-processing shown in box 5 in Fig. \ref{fig:overall} after the weighted sum in box 6, so that it is learned for the combined seasonal and trend components. The variant is referred to as \textit{TimeFlex-joint post-processing} (\textit{TimeFlex\_jpp}).

\section{Experimental Setup}
To ensure comparability, the experimental setup closely follows the methodologies described in \cite{TimeMixer} and \cite{wang2024survey}. In contrast to those surveys, we apply the same splitting rate for training, validation, and test data in all trainings. The full implementation as well as all parameter settings the datasets and additional hyperparameters can be found in the corresponding repository\footnote{The evaluation of the models is structured according to their architectural characteristics. Comprehensive results of all experiments on the benchmarking datasets are available in the repository. The results for the sampled test data sets are summarized within Tables \ref{tab:local_period}, \ref{tab:period}, \ref{tab:rational}, \ref{tab:combined} and \ref{tab:se} showing Mean Squared Error (MSE) and Mean Absolute Error (MAE)). The best results on the test data are highlighted as \colorbox[RGB]{31,175,34}{\textcolor{white}{\textbf{bold-white text}}} second best as \colorbox[RGB]{31,175,34}{\textcolor{black}{\textbf{bold-black}}}. The table further employs a color-coded scale for shading of the cells, where \colorbox[RGB]{31,175,34}{{green}} denotes the best values, white represents average values, and \colorbox[RGB]{255,0,0}{{red}} indicates the worst values. In the following, we will briefly present individual results and an evaluation for specific models, individual characteristics, and how specific parts address these. This will summarize the main findings.}. All models are trained on the same machine\footnote{Ubuntu 22.04, x86-64, an Intel Xeon W5-2465X processor, offering 16 physical cores, 32 logical CPUs and an NVIDIA RTX A6000 with 48 GB of GDDR6 memory using CUDA Version 12.2}. Further settings are listed in Table \ref{tab:exp}. 
\begin{table}[H]
\centering
\resizebox{\columnwidth}{!}{
\begin{tabular}{|l|l|}
\hline
\textbf{Parameter} & \textbf{Value} \\
\hline
Input sequence length & 96 \\
Prediction sequence lengths & \{96, 192, 336, 720\} \\
Dataset split (Train/Val/Test) & 70\% / 20\% / 10\% \\
Optimizer, Scaler & Adam, Standard \\
Discard incomplete batches & True \\
\hline
\textbf{Scheduler (Reduce on Plateau)} & Patience: 2, Factor: 0.1, Mode: min \\
\hline
\textbf{Early Stopping} & Patience: 10, Delta: 0.00001 \\
\hline
\end{tabular}
}
\caption{Experimental Set-Up} \label{tab:exp}
\end{table}
The models are compared by training them on our generated dataset, as described in Section \ref{sec:data_gen} and on common benchmarking datasets as provided by \cite{TimeMixer}, i.e. the ETT datasets (including ETTh1, ETTh2, ETTm1 and ETTm2), Weather, Traffic, Electricity and Exchange Rate\footnote{When training with an input sequence length of 92 steps and forecasting 720 steps, the total sequence length is 812. The Exchange Rate data set has a total length of 7589 steps, such that the number of complete sequences is 9. Due to the small number of training data, we omit forecasting 720 steps for the Exchange Rate dataset.
}.

\section{Evaluation}

The evaluation of the models is structured according to their architectural characteristics. Comprehensive results of all experiments on the benchmarking datasets are also available on Github. The results for the sampled test datasets are summarized within tables \ref{tab:local_period}, \ref{tab:period}, \ref{tab:rational}, \ref{tab:combined} and \ref{tab:se} showing the Mean Squared Error (MSE) and the Mean Absolute Error (MAE)). The best results on the test data are highlighted as \colorbox[RGB]{31,175,34}{\textcolor{white}{\textbf{bold-white text}}} second best as \colorbox[RGB]{31,175,34}{\textcolor{black}{\textbf{bold-black}}}. The table also uses a color-coded scale for cell shading, where \colorbox[RGB]{31,175,34}{{green}} denotes the best values, white represents the average values, and \colorbox[RGB]{255,0,0}{{red}} indicates the worst values. We will now briefly present the results and evaluate how specific models and characteristics address individual aspects.

\input{tables/sampled/locally_periodic_kernel_table}
\input{tables/sampled/periodic_kernel_table}
\input{tables/sampled/rational_quadratic_kernel_table}
\input{tables/sampled/combined_kernel_table}
\input{tables/sampled/se_kernel_table}
\input{tables/bench/weather_table}
\input{tables/bench/Etth1_table}
\input{tables/bench/Ettm2_table}

\subsection{Linear Dense Layer}
A central element of \textit{RLinear}, \textit{NLinear}, \textit{TiDE} and \textit{SparseTSF} is a fully connected linear layer. This elementary architecture yields relatively good results on the datasets sampled with a periodic (see Table \ref{tab:period}) and a SE-kernel (see Table \ref{tab:se}) that have no underlying trend since single reoccurring pattern can be represented by linear mapping of input and output. For datasets with complex patterns or pronounced trends, these models show comparatively inferior results.

\subsection{Decomposition}
Comparing methods that incorporate a decomposition block with similar models lacking a decomposition, like \textit{DLinear} with \textit{RLinear} and \textit{NLinear} or \textit{TimeFlex} with \textit{WaveNet}, we observe that those with a decomposition block consistently demonstrate superior performance. This holds particularly true for the locally periodic dataset (see Table \ref{tab:local_period}), the rational-quadratic (see Table \ref{tab:rational}) and combined datasets (see Table \ref{tab:combined}) exhibiting not only simple periodicity but also underlying trends or more intricate recurring patterns. The same can be demonstrated on the weather dataset--as a real world example--which has long and short-term trends as well as seasonal patterns (see Table \ref{tab:weather}). The best performing models are \textit{CycleNet} using a known cycle length to explicitly model periodic patterns and subtract it, such that only the residual part is learned and our \textit{TimeFlex} model which successfully learns the trend and seasonal part separately within in the time and frequency domain, respectively.

\subsection{Fourier and Wavelet transform}

When transforming a sequence from the time domain to the frequency domain, information about changes over time, such as trends, is lost. However, the frequency representation eases the identification of patterns that recur at regular time intervals. As previously described, simple model architectures with a linear dense layer like \textit{D-/R-/N-Linear} are capable of learning periodicity within data if it does not exhibit further characteristics and patterns (Table \ref{tab:period} and \ref{tab:se}). When applying these simple model architecture with just a linear dense layer to the locally periodic dataset (see Table \ref{tab:local_period}) they produce inferior outcomes being challenged when learning the periodic patterns despite the inherent trend. Comparing \textit{WaveMask/WaveMix}  with \textit{DLinear}, the application of Wavelet transforms yields better results. The same observations can be made on the ETT (see Tables \ref{tab:ETTh1}, \ref{tab:ETTm2}) and Exchange Rate datasets, which are characterized by complex periodic patterns and trends. In our ablation studies, we trained \textit{TimeFlex} with and without transforming the seasonal part using 1- or 2-D FFT and the models with FFT performed better among almost all datasets. Applying the FFT only to the seasonal part, the model does not suffer a loss of timely information on long-term changes while improving the learning of periodic patterns through processing them in the frequency domain.

\subsection{Computational and Model Complexity}

\begin{figure}
    \centering
    \includegraphics[width=1\linewidth, trim=1cm 1cm 1cm 1cm, clip]{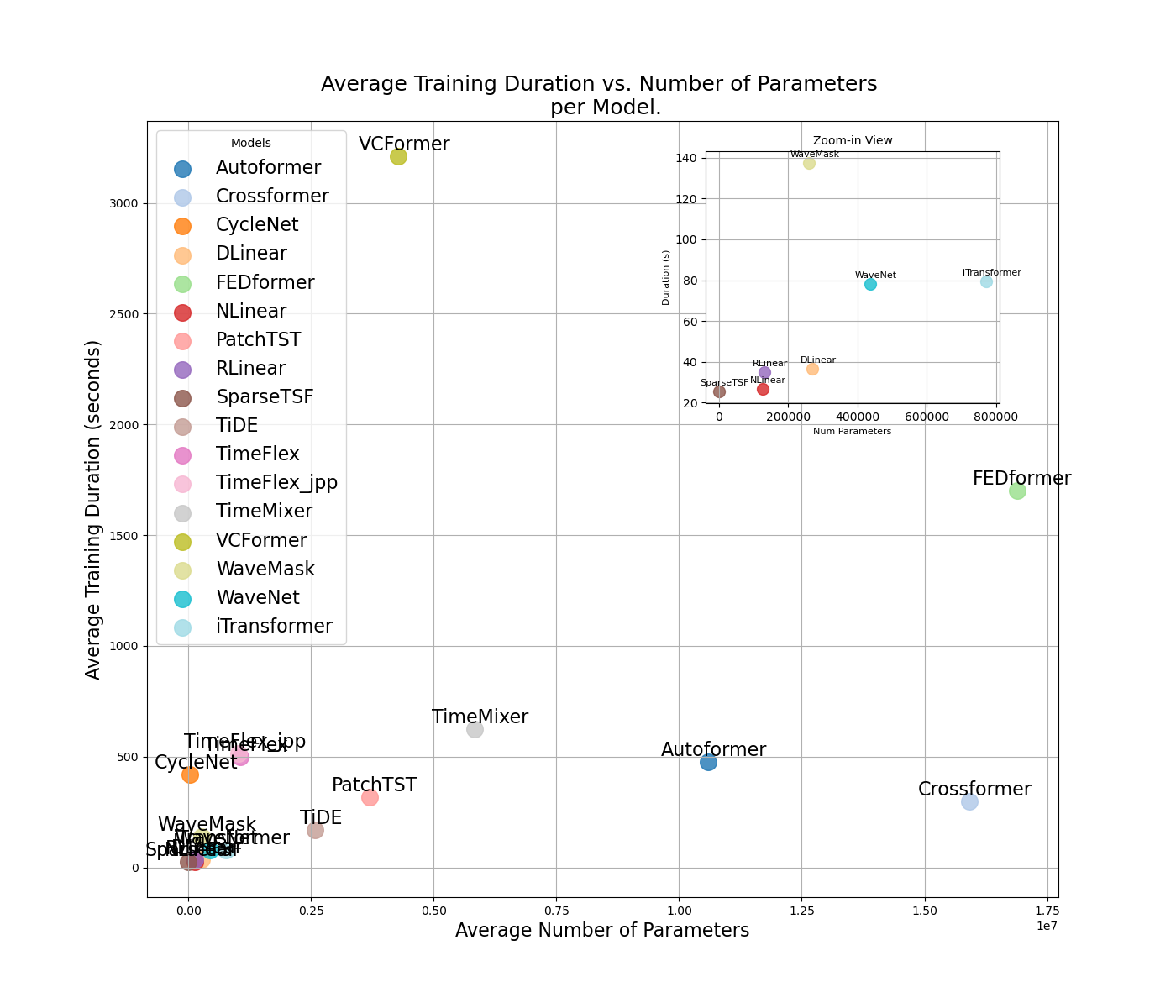}
    \caption{Average runtime over number of parameters.}
    \label{fig:avg run}
\end{figure}

Computational complexity does scale differently with respect to the number of learnable parameters for the different models and there is a significant effect on computation time depending on model type as well as model size.
In Figure \ref{fig:avg run} the computation time of training the different models is plotted over the number of parameters.

\textit{SparseTSF} outperforms many models on all considered datasets while having a small number of parameters and a low computational complexity. But, for the datasets with more complex patterns \textit{SparseTSF} fails to capture them, i.e. the model is underfitting. \textit{CycleNet}, \textit{WaveMask} and \textit{TimeFlex} achieve best performance on most datasets while utilizing few parameters and requiring significantly less computation time compared to Transformers. \textit{Crossformer} is a very efficient Transformer architecture given its large number of parameters and considering the 2-dimensional attention mechanism, standing out as the best overall Transformer model, except on the periodic dataset. \textit{iTransformer} demonstrates exceptionally fast training times, surpassing non-Transformer architectures such as WaveMask. 
The Transformer architectures generally have longer training times compared to other architecture types, but they are still outperformed by the best-performing MLP- and CNN-based architectures.

\subsection{Ablation study on \textit{TimeFlex}.} \label{A2}

\begin{figure}
    \centering
    \includegraphics[width=0.9\linewidth, trim=0cm 0cm 0cm 2cm, clip]{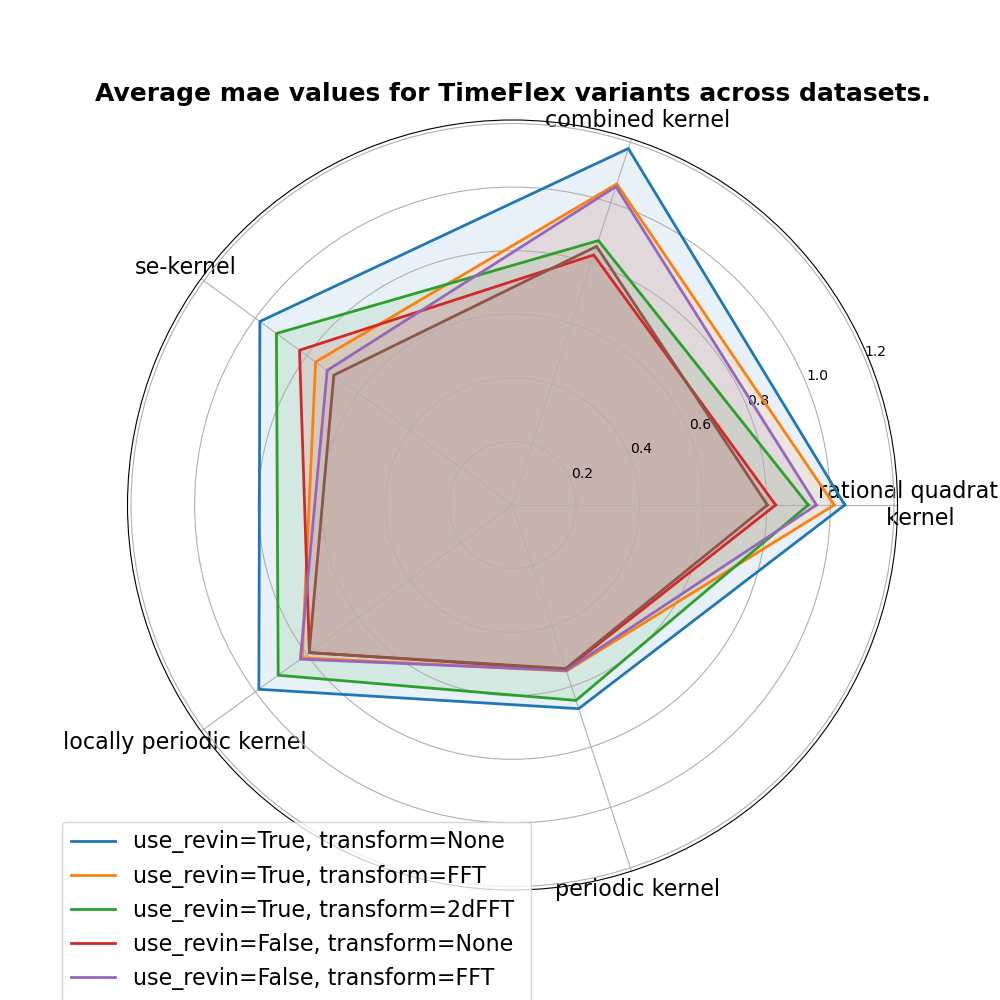}
    \caption{MAE of \textit{TimeFlex} with/without RevIn, 1D- and 2D FFT.}
    \label{fig:abl}
\end{figure}

\begin{figure}
    \centering
    \includegraphics[width=0.9\linewidth, trim=0cm 0cm 0cm 2cm, clip]{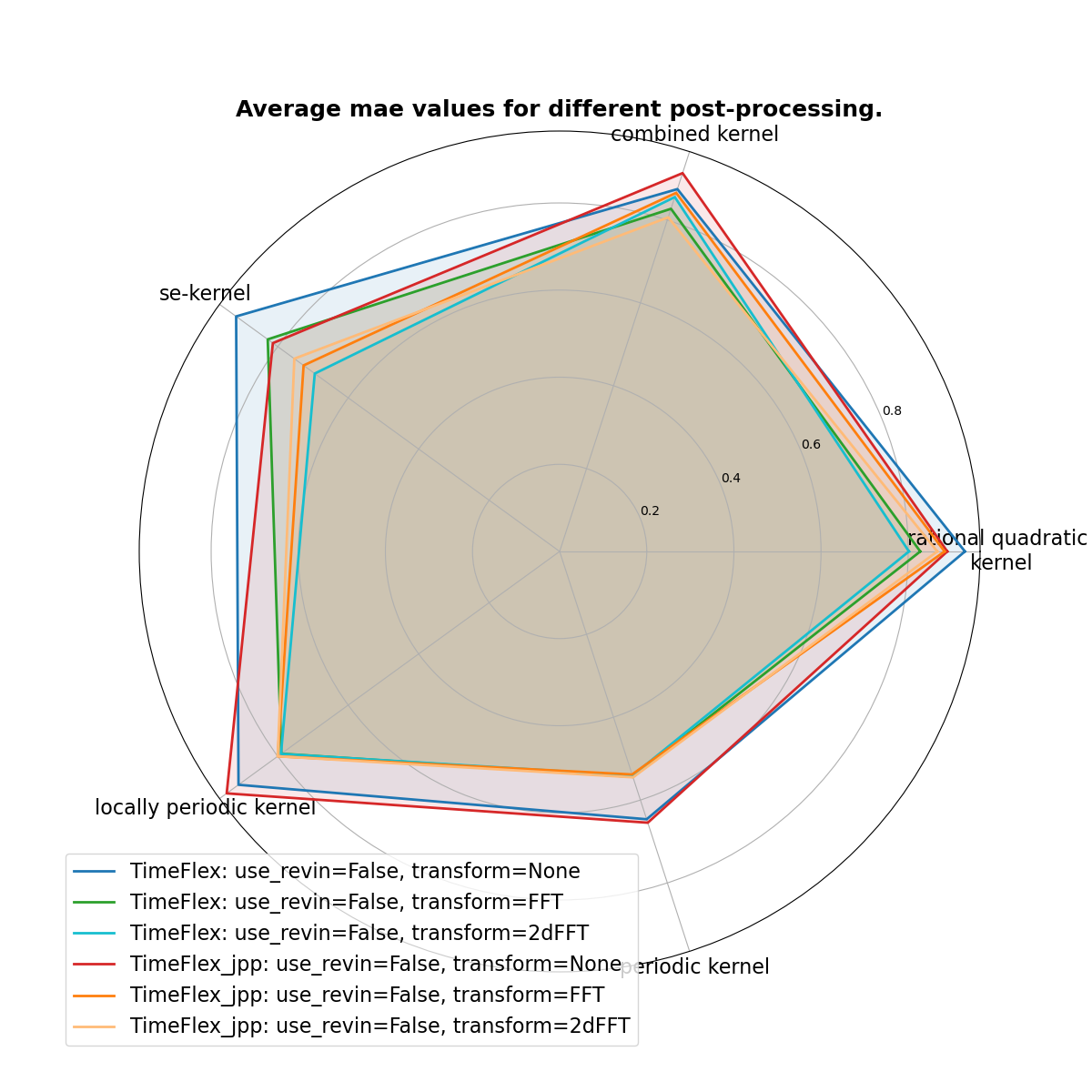}
    \caption{Comparison of \textit{TimeFlex} with \textit{TimeFlex\_jpp}, i.e. post-processing applied after the weighted sum.}
    \label{fig:jpp}
\end{figure}

The results of different \textit{TimeFlex} variants on the sampled datasets are depicted in Figure \ref{fig:abl}. All experimental results in this paper are obtained using the CI-variant, which outperforms the CD-variant in any setting. Applying RevIN within \textit{TimeFlex} yields inferior results among most experiments. Applying RevIn to smooth and stabilize the data before applying decomposition can reduce the decomposition accuracy. \textit{TimeFlex} with a 2D Fourier transformation excels on all datasets other than the one sampled with the combined kernel. Remarkably, applying FFT across the time and variate domain gives better results than the standard 1D-FFT since the features in the sampled datasets are provably correlated and it enables an information exchange among features. We further compared whether to apply the post-processing on the seasonal part before the weighted sum (blue and green) or after the sum (red and orange). The results are illustrated in Figure \ref{fig:jpp} (results with RevIN are similar and left aside for visualization purposes). For most settings and datasets the \textit{TimeFlex} variant with the post-processing block before the weighted sum (as shown in Figure \ref{fig:overall}) shows better results. Applying a joint post-processing on season and trend is only superior for SE and rational quadratic kernel datasets when no FFT is applied. For the 1D-FFT it is slightly better on the SE-kernel dataset and for the 2D-FFT on the combined kernel dataset. 

\subsection{Further observations}
Comparing models applying RevIN with those applying Batch Normalization on different datasets, the following observations were made: \textit{RLinear} performs better on the datasets ETTh1, ETTm1, ETTm2 (see Tables \ref{tab:ETTh1}, \ref{tab:ETTm2} and Traffic, while \textit{NLinear} is better on ETTh2 and Exchange Rate. For the sampled datasets \textit{NLinear} performs better on the comparably smooth SE-kernel dataset (Table \ref{tab:se}, while \textit{RLinear} yields better results for the less smooth datasets sampled with a rational quadratic and combined kernel (Tables \ref{tab:rational} and \ref{tab:combined}). Thus, the experimental results on the sampled data sets align well with expectations based on the earlier explanation. The ablation study in \cite{Cyclenet} with and without RevIN also shows that \textit{CycleNet} benefits from Instance Normalization if the data exhibits a distribution shift. RevIN normalizes each sequence independently, which is beneficial for highly non-stationary time series, but might lead to oversmoothing of more consistent ones. Like Batch Normalization RevIN is not robust to varying sequence lengths and might suffer a loss of global context. 
To draw conclusions about the long-term and short-term forecasting performance of different models we compared their performance for the smaller (96, 192 steps) and the longer (336, 720 steps) forecasting horizons. In our experimental results, we did not observe a clear superiority of any model either on the short-term forecasting or on the long-term forecasting. The performance of models within our experiments depends more on the specific characteristics of the data than on the prediction lengths.  On small and less complex datasets, transformers cannot fully leverage their strengths and tend to overfit. Whether they perform better on very large datasets and time series with complex patterns remains an open question for future work.

\section{Conclusion}
In this article, we generated datasets using Gaussian Processes (GPs) that allow for extensive evaluations of time series forecasting models, considering specific architectural choices and data attributes. This targeted approach allowed us to systematically analyze how different architectures handle specific time series characteristics. Our findings indicate that separate processing of seasonal and trend components yields significantly better results for non-stationary datasets. Additionally, the use of Fourier and Wavelet transforms substantially improves model performance on data with complex periodic patterns, particularly when only applied to the seasonal part. Building on these insights, we proposed \textit{TimeFlex} as a novel time series forecasting architecture that decomposes seasonal and trend components, enhancing the ability to learn periodic patterns by processing the seasonal part in the frequency domain and effectively learning trends over time in the time domain. The model achieves state-of-the-art performance compared to other deep learning models without requiring prior domain knowledge. Among the evaluated models, \textit{SparseTSF} showed to offer a good balance between complexity and accuracy, while models with more parameters like \textit{TimeFlex}, \textit{CycleNet}, or \textit{WaveMask} excel in terms of accuracy. However, more sophisticated Transformers with a larger number of parameters did not show leading performance, potentially due to overfitting on the relatively small benchmark datasets. Future work should evaluate these Transformer architectures on larger datasets to fully assess their capabilities and limitations.

\bibliographystyle{IEEEtran}
\bibliography{references}

\end{document}

%% file: tables/sampled/locally_periodic_kernel_table.tex
\begin{table}[htbp]
\centering
\resizebox{\columnwidth}{!}{
\begin{tabular}{|l|rr|rr|rr|rr|}
\hline
$\empty$ & \multicolumn{2}{c|}{\textbf{96 steps}} & \multicolumn{2}{c|}{\textbf{196 steps}} & \multicolumn{2}{c|}{\textbf{336 steps}} & \multicolumn{2}{c|}{\textbf{720 steps}} \\
\hline
\textbf{Model} & \textbf{MSE} & \textbf{MAE} & \textbf{MSE} & \textbf{MAE} & \textbf{MSE} & \textbf{MAE} & \textbf{MSE} & \textbf{MAE} \\
\hline
CycleNet & \cellcolor[RGB]{31,175,34}\textbf{\textcolor{white}{1.0196}} & \cellcolor[RGB]{31,175,34}\textbf{\textcolor{white}{0.7952}} & \cellcolor[RGB]{31,175,34}\textbf{\textcolor{white}{0.9422}} & \cellcolor[RGB]{31,175,34}\textbf{\textcolor{white}{0.7678}} & \cellcolor[RGB]{31,175,34}\textbf{\textcolor{white}{0.9460}} & \cellcolor[RGB]{31,175,34}\textbf{\textcolor{white}{0.7734}} & \cellcolor[RGB]{48,181,51}\textbf{0.9969} & \cellcolor[RGB]{57,184,59}\textbf{0.8006} \\
\hline
WaveMask & \cellcolor[RGB]{52,182,55}\textbf{1.0430} & \cellcolor[RGB]{45,180,48}\textbf{0.8004} & \cellcolor[RGB]{33,176,36}\textbf{0.9448} & \cellcolor[RGB]{31,175,34}\textbf{0.7681} & \cellcolor[RGB]{44,179,47}0.9566 & \cellcolor[RGB]{42,179,45}0.7769 & \cellcolor[RGB]{56,184,59}1.0024 & \cellcolor[RGB]{59,185,61}0.8012 \\
\hline
DLinear & \cellcolor[RGB]{56,184,59}1.0468 & \cellcolor[RGB]{49,181,52}0.8018 & \cellcolor[RGB]{41,178,44}0.9516 & \cellcolor[RGB]{36,177,39}0.7696 & \cellcolor[RGB]{52,182,54}0.9628 & \cellcolor[RGB]{48,181,51}0.7788 & \cellcolor[RGB]{63,186,66}1.0075 & \cellcolor[RGB]{67,188,70}0.8034 \\
\hline
RLinear & \cellcolor[RGB]{141,214,143}1.1373 & \cellcolor[RGB]{156,219,158}0.8396 & \cellcolor[RGB]{153,218,155}1.0498 & \cellcolor[RGB]{194,233,194}0.8182 & \cellcolor[RGB]{112,204,114}1.0103 & \cellcolor[RGB]{116,205,118}0.7997 & \cellcolor[RGB]{150,217,151}1.0668 & \cellcolor[RGB]{153,218,155}0.8253 \\
\hline
NLinear & \cellcolor[RGB]{134,211,135}1.1294 & \cellcolor[RGB]{136,212,138}0.8326 & \cellcolor[RGB]{159,221,161}1.0552 & \cellcolor[RGB]{196,234,197}0.8189 & \cellcolor[RGB]{117,205,119}1.0142 & \cellcolor[RGB]{114,204,116}0.7991 & \cellcolor[RGB]{162,222,164}1.0756 & \cellcolor[RGB]{162,222,164}0.8277 \\
\hline
Crossformer & \cellcolor[RGB]{69,188,72}1.0609 & \cellcolor[RGB]{78,192,81}0.8121 & \cellcolor[RGB]{44,179,47}0.9541 & \cellcolor[RGB]{43,179,46}0.7716 & \cellcolor[RGB]{55,183,58}0.9654 & \cellcolor[RGB]{52,182,55}0.7800 & \cellcolor[RGB]{57,184,60}1.0034 & \cellcolor[RGB]{61,185,63}0.8017 \\
\hline
TiDE & \cellcolor[RGB]{123,207,124}1.1176 & \cellcolor[RGB]{147,216,148}0.8362 & \cellcolor[RGB]{146,216,148}1.0437 & \cellcolor[RGB]{190,232,191}0.8172 & \cellcolor[RGB]{111,203,113}1.0100 & \cellcolor[RGB]{112,204,114}0.7986 & \cellcolor[RGB]{147,216,149}1.0652 & \cellcolor[RGB]{153,218,154}0.8253 \\
\hline
SparseTSF & \cellcolor[RGB]{183,229,184}1.1823 & \cellcolor[RGB]{233,247,233}0.8666 & \cellcolor[RGB]{171,225,172}1.0652 & \cellcolor[RGB]{221,243,222}0.8267 & \cellcolor[RGB]{113,204,115}1.0109 & \cellcolor[RGB]{115,205,117}0.7995 & \cellcolor[RGB]{146,216,148}1.0645 & \cellcolor[RGB]{151,217,152}0.8247 \\
\hline
PatchTST & \cellcolor[RGB]{189,231,190}1.1889 & \cellcolor[RGB]{234,247,234}0.8671 & \cellcolor[RGB]{164,222,165}1.0590 & \cellcolor[RGB]{218,241,218}0.8255 & \cellcolor[RGB]{109,202,111}1.0079 & \cellcolor[RGB]{110,203,112}0.7981 & \cellcolor[RGB]{141,214,143}1.0611 & \cellcolor[RGB]{147,216,148}0.8237 \\
\hline
VCFormer & \cellcolor[RGB]{192,232,192}1.1911 & \cellcolor[RGB]{238,249,238}0.8683 & \cellcolor[RGB]{178,227,179}1.0718 & \cellcolor[RGB]{227,245,227}0.8283 & \cellcolor[RGB]{113,204,115}1.0115 & \cellcolor[RGB]{113,204,115}0.7990 & \cellcolor[RGB]{142,214,144}1.0617 & \cellcolor[RGB]{148,216,149}0.8240 \\
\hline
Autoformer & \cellcolor[RGB]{192,232,193}1.1914 & \cellcolor[RGB]{242,250,243}0.8700 & \cellcolor[RGB]{168,224,169}1.0630 & \cellcolor[RGB]{221,242,221}0.8265 & \cellcolor[RGB]{107,202,109}1.0065 & \cellcolor[RGB]{110,203,112}0.7979 & \cellcolor[RGB]{139,213,140}1.0592 & \cellcolor[RGB]{152,218,153}0.8250 \\
\hline
TimeMixer & \cellcolor[RGB]{99,199,101}1.0926 & \cellcolor[RGB]{92,196,94}0.8169 & \cellcolor[RGB]{146,216,147}1.0435 & \cellcolor[RGB]{198,234,199}0.8194 & \cellcolor[RGB]{106,201,108}1.0053 & \cellcolor[RGB]{100,199,102}0.7949 & \cellcolor[RGB]{141,214,142}1.0608 & \cellcolor[RGB]{147,216,148}0.8236 \\
\hline
WaveNet & \cellcolor[RGB]{255,44,44}1.4546 & \cellcolor[RGB]{255,0,0}0.9533 & \cellcolor[RGB]{255,108,108}1.2517 & \cellcolor[RGB]{255,70,70}0.8869 & \cellcolor[RGB]{255,0,0}1.2997 & \cellcolor[RGB]{255,0,0}0.9119 & \cellcolor[RGB]{255,0,0}1.2932 & \cellcolor[RGB]{255,0,0}0.9084 \\
\hline
FEDformer & \cellcolor[RGB]{255,0,0}1.4967 & \cellcolor[RGB]{255,48,48}0.9383 & \cellcolor[RGB]{255,0,0}1.3350 & \cellcolor[RGB]{255,0,0}0.9061 & \cellcolor[RGB]{255,251,251}1.1252 & \cellcolor[RGB]{197,234,198}0.8249 & \cellcolor[RGB]{255,136,136}1.2108 & \cellcolor[RGB]{255,118,118}0.8819 \\
\hline
iTransformer & \cellcolor[RGB]{189,231,190}1.1888 & \cellcolor[RGB]{235,247,235}0.8672 & \cellcolor[RGB]{174,226,175}1.0682 & \cellcolor[RGB]{224,243,224}0.8274 & \cellcolor[RGB]{113,204,115}1.0113 & \cellcolor[RGB]{113,204,115}0.7988 & \cellcolor[RGB]{137,212,138}1.0578 & \cellcolor[RGB]{143,215,145}0.8228 \\
\hline
TimeFlex & \cellcolor[RGB]{77,191,79}1.0687 & \cellcolor[RGB]{81,192,83}0.8129 & \cellcolor[RGB]{47,180,50}0.9569 & \cellcolor[RGB]{41,178,44}0.7712 & \cellcolor[RGB]{34,176,37}\textbf{0.9486} & \cellcolor[RGB]{39,177,42}\textbf{0.7759} & \cellcolor[RGB]{31,175,34}\textbf{\textcolor{white}{0.9848}} & \cellcolor[RGB]{31,175,34}\textbf{\textcolor{white}{0.7940}} \\
\hline
\end{tabular}
}
\caption{Full results (MSE and MAE) of different models trained on the dataset with locally periodic kernel.} \label{tab:local_period}
\end{table}

%% file: tables/sampled/periodic_kernel_table.tex
\begin{table}[htbp]
\centering
\resizebox{\columnwidth}{!}{
\begin{tabular}{|l|rr|rr|rr|rr|}
\hline
$\empty$ & \multicolumn{2}{c|}{\textbf{96 steps}} & \multicolumn{2}{c|}{\textbf{196 steps}} & \multicolumn{2}{c|}{\textbf{336 steps}} & \multicolumn{2}{c|}{\textbf{720 steps}} \\
\hline
\textbf{Model} & \textbf{MSE} & \textbf{MAE} & \textbf{MSE} & \textbf{MAE} & \textbf{MSE} & \textbf{MAE} & \textbf{MSE} & \textbf{MAE} \\
\hline
CycleNet & \cellcolor[RGB]{31,175,34}\textbf{\textcolor{white}{0.4520}} & \cellcolor[RGB]{31,175,34}\textbf{\textcolor{white}{0.5276}} & \cellcolor[RGB]{31,175,34}\textbf{\textcolor{white}{0.4717}} & \cellcolor[RGB]{31,175,34}\textbf{\textcolor{white}{0.5406}} & \cellcolor[RGB]{31,175,34}\textbf{\textcolor{white}{0.4630}} & \cellcolor[RGB]{31,175,34}\textbf{\textcolor{white}{0.5347}} & \cellcolor[RGB]{31,175,34}\textbf{\textcolor{white}{0.4599}} & \cellcolor[RGB]{31,175,34}\textbf{\textcolor{white}{0.5372}} \\
\hline
WaveMask & \cellcolor[RGB]{63,186,65}0.4940 & \cellcolor[RGB]{73,190,76}0.5535 & \cellcolor[RGB]{52,182,55}0.5016 & \cellcolor[RGB]{55,183,58}0.5570 & \cellcolor[RGB]{52,182,55}0.4904 & \cellcolor[RGB]{55,183,58}0.5498 & \cellcolor[RGB]{62,186,65}0.4995 & \cellcolor[RGB]{66,187,69}0.5588 \\
\hline
DLinear & \cellcolor[RGB]{64,186,67}0.4958 & \cellcolor[RGB]{73,190,76}0.5535 & \cellcolor[RGB]{53,182,55}0.5025 & \cellcolor[RGB]{55,183,58}0.5568 & \cellcolor[RGB]{52,182,55}0.4902 & \cellcolor[RGB]{55,183,58}0.5498 & \cellcolor[RGB]{64,186,66}0.5009 & \cellcolor[RGB]{68,188,70}0.5598 \\
\hline
RLinear & \cellcolor[RGB]{62,186,64}0.4927 & \cellcolor[RGB]{74,190,77}0.5542 & \cellcolor[RGB]{56,183,58}0.5069 & \cellcolor[RGB]{63,186,65}0.5621 & \cellcolor[RGB]{58,184,60}0.4972 & \cellcolor[RGB]{65,187,67}0.5557 & \cellcolor[RGB]{65,187,67}0.5022 & \cellcolor[RGB]{71,189,73}0.5615 \\
\hline
NLinear & \cellcolor[RGB]{67,187,69}0.4992 & \cellcolor[RGB]{81,193,84}0.5584 & \cellcolor[RGB]{56,184,59}0.5073 & \cellcolor[RGB]{62,186,65}0.5618 & \cellcolor[RGB]{59,185,62}0.4987 & \cellcolor[RGB]{65,187,67}0.5558 & \cellcolor[RGB]{68,188,70}0.5061 & \cellcolor[RGB]{74,190,76}0.5635 \\
\hline
Crossformer & \cellcolor[RGB]{204,236,204}0.6790 & \cellcolor[RGB]{229,245,230}0.6481 & \cellcolor[RGB]{249,252,249}0.7766 & \cellcolor[RGB]{255,243,243}0.6966 & \cellcolor[RGB]{223,243,224}0.7041 & \cellcolor[RGB]{235,248,236}0.6609 & \cellcolor[RGB]{198,234,199}0.6685 & \cellcolor[RGB]{200,235,201}0.6406 \\
\hline
TiDE & \cellcolor[RGB]{53,182,55}0.4810 & \cellcolor[RGB]{62,186,64}0.5467 & \cellcolor[RGB]{52,182,54}0.5013 & \cellcolor[RGB]{56,184,59}0.5578 & \cellcolor[RGB]{55,183,58}0.4941 & \cellcolor[RGB]{61,185,63}0.5533 & \cellcolor[RGB]{59,185,61}0.4950 & \cellcolor[RGB]{63,186,66}0.5571 \\
\hline
SparseTSF & \cellcolor[RGB]{59,185,62}0.4893 & \cellcolor[RGB]{66,187,68}0.5490 & \cellcolor[RGB]{55,183,58}0.5059 & \cellcolor[RGB]{61,185,64}0.5610 & \cellcolor[RGB]{58,184,61}0.4979 & \cellcolor[RGB]{65,187,67}0.5559 & \cellcolor[RGB]{62,186,65}0.4991 & \cellcolor[RGB]{68,188,70}0.5597 \\
\hline
PatchTST & \cellcolor[RGB]{48,181,51}0.4752 & \cellcolor[RGB]{55,183,58}0.5425 & \cellcolor[RGB]{50,182,53}0.4991 & \cellcolor[RGB]{58,184,61}0.5591 & \cellcolor[RGB]{45,180,48}0.4815 & \cellcolor[RGB]{53,182,55}0.5484 & \cellcolor[RGB]{46,180,49}0.4790 & \cellcolor[RGB]{50,181,53}0.5490 \\
\hline
VCFormer & \cellcolor[RGB]{255,0,0}1.0393 & \cellcolor[RGB]{255,0,0}0.7992 & \cellcolor[RGB]{255,0,0}1.0980 & \cellcolor[RGB]{255,0,0}0.8390 & \cellcolor[RGB]{255,0,0}1.0235 & \cellcolor[RGB]{255,0,0}0.8107 & \cellcolor[RGB]{255,0,0}1.0168 & \cellcolor[RGB]{255,0,0}0.8099 \\
\hline
Autoformer & \cellcolor[RGB]{255,63,63}0.9662 & \cellcolor[RGB]{255,38,38}0.7787 & \cellcolor[RGB]{34,176,37}\textbf{0.4761} & \cellcolor[RGB]{31,175,34}\textbf{0.5407} & \cellcolor[RGB]{133,211,135}0.5913 & \cellcolor[RGB]{134,212,136}0.5987 & \cellcolor[RGB]{255,175,175}0.8250 & \cellcolor[RGB]{255,167,167}0.7201 \\
\hline
TimeMixer & \cellcolor[RGB]{43,179,46}0.4686 & \cellcolor[RGB]{53,182,55}0.5410 & \cellcolor[RGB]{42,179,45}0.4875 & \cellcolor[RGB]{47,180,50}0.5516 & \cellcolor[RGB]{47,180,50}0.4839 & \cellcolor[RGB]{50,181,53}0.5466 & \cellcolor[RGB]{44,179,46}0.4761 & \cellcolor[RGB]{46,180,49}0.5469 \\
\hline
WaveNet & \cellcolor[RGB]{233,247,233}0.7171 & \cellcolor[RGB]{249,253,249}0.6601 & \cellcolor[RGB]{187,230,188}0.6907 & \cellcolor[RGB]{196,234,197}0.6508 & \cellcolor[RGB]{223,243,224}0.7042 & \cellcolor[RGB]{244,251,244}0.6663 & \cellcolor[RGB]{228,245,229}0.7057 & \cellcolor[RGB]{238,248,238}0.6632 \\
\hline
FEDformer & \cellcolor[RGB]{86,194,89}0.5253 & \cellcolor[RGB]{100,199,102}0.5696 & \cellcolor[RGB]{114,204,116}0.5881 & \cellcolor[RGB]{120,206,122}0.6002 & \cellcolor[RGB]{93,197,95}0.5410 & \cellcolor[RGB]{101,200,103}0.5780 & \cellcolor[RGB]{68,188,70}0.5059 & \cellcolor[RGB]{74,190,76}0.5635 \\
\hline
iTransformer & \cellcolor[RGB]{155,219,156}0.6150 & \cellcolor[RGB]{173,225,174}0.6139 & \cellcolor[RGB]{134,212,136}0.6167 & \cellcolor[RGB]{158,220,160}0.6258 & \cellcolor[RGB]{94,197,96}0.5428 & \cellcolor[RGB]{116,205,118}0.5876 & \cellcolor[RGB]{128,209,129}0.5805 & \cellcolor[RGB]{151,218,152}0.6105 \\
\hline
TimeFlex & \cellcolor[RGB]{37,177,40}\textbf{0.4605} & \cellcolor[RGB]{38,177,41}\textbf{0.5323} & \cellcolor[RGB]{36,176,39}0.4789 & \cellcolor[RGB]{37,177,40}0.5449 & \cellcolor[RGB]{35,176,38}\textbf{0.4683} & \cellcolor[RGB]{35,176,38}\textbf{0.5376} & \cellcolor[RGB]{32,175,35}\textbf{0.4620} & \cellcolor[RGB]{32,175,35}\textbf{0.5383} \\
\hline
\end{tabular}
}
\caption{Full results (MSE and MAE) of different models trained on the dataset with periodic kernel.} \label{tab:period}
\end{table}

%% file: tables/sampled/rational_quadratic_kernel_table.tex
\begin{table}[htbp]
\centering
\resizebox{\columnwidth}{!}{
\begin{tabular}{|l|rr|rr|rr|rr|}
\hline
$\empty$ & \multicolumn{2}{c|}{\textbf{96 steps}} & \multicolumn{2}{c|}{\textbf{196 steps}} & \multicolumn{2}{c|}{\textbf{336 steps}} & \multicolumn{2}{c|}{\textbf{720 steps}} \\
\hline
\textbf{Model} & \textbf{MSE} & \textbf{MAE} & \textbf{MSE} & \textbf{MAE} & \textbf{MSE} & \textbf{MAE} & \textbf{MSE} & \textbf{MAE} \\
\hline
CycleNet & \cellcolor[RGB]{31,175,34}\textbf{\textcolor{white}{1.0214}} & \cellcolor[RGB]{31,175,34}\textbf{\textcolor{white}{0.8023}} & \cellcolor[RGB]{43,179,46}0.9650 & \cellcolor[RGB]{62,186,65}0.7825 & \cellcolor[RGB]{39,178,42}1.0852 & \cellcolor[RGB]{56,184,59}0.8229 & \cellcolor[RGB]{31,175,34}\textbf{\textcolor{white}{0.9751}} & \cellcolor[RGB]{31,175,34}\textbf{\textcolor{white}{0.7753}} \\
\hline
WaveMask & \cellcolor[RGB]{45,180,48}1.0566 & \cellcolor[RGB]{48,181,51}0.8142 & \cellcolor[RGB]{33,176,36}\textbf{0.9445} & \cellcolor[RGB]{35,176,38}0.7623 & \cellcolor[RGB]{31,175,34}\textbf{\textcolor{white}{1.0701}} & \cellcolor[RGB]{31,175,34}\textbf{\textcolor{white}{0.8069}} & \cellcolor[RGB]{51,182,54}\textbf{0.9961} & \cellcolor[RGB]{35,176,38}\textbf{0.7771} \\
\hline
DLinear & \cellcolor[RGB]{49,181,51}1.0644 & \cellcolor[RGB]{49,181,52}0.8152 & \cellcolor[RGB]{42,179,45}0.9625 & \cellcolor[RGB]{46,180,48}0.7703 & \cellcolor[RGB]{35,176,38}\textbf{1.0773} & \cellcolor[RGB]{32,175,35}\textbf{0.8078} & \cellcolor[RGB]{55,183,57}0.9994 & \cellcolor[RGB]{40,178,43}0.7791 \\
\hline
RLinear & \cellcolor[RGB]{113,204,114}1.2157 & \cellcolor[RGB]{213,240,214}0.9279 & \cellcolor[RGB]{121,207,123}1.1325 & \cellcolor[RGB]{199,235,200}0.8809 & \cellcolor[RGB]{191,232,192}1.3385 & \cellcolor[RGB]{239,249,239}0.9365 & \cellcolor[RGB]{255,165,165}1.2787 & \cellcolor[RGB]{255,118,118}0.9065 \\
\hline
NLinear & \cellcolor[RGB]{155,219,156}1.3163 & \cellcolor[RGB]{246,251,246}0.9503 & \cellcolor[RGB]{136,212,138}1.1649 & \cellcolor[RGB]{202,236,202}0.8830 & \cellcolor[RGB]{203,236,203}1.3587 & \cellcolor[RGB]{242,250,242}0.9381 & \cellcolor[RGB]{255,101,101}1.3352 & \cellcolor[RGB]{255,49,49}0.9294 \\
\hline
Crossformer & \cellcolor[RGB]{139,213,141}1.2790 & \cellcolor[RGB]{255,236,236}0.9671 & \cellcolor[RGB]{31,175,34}\textbf{\textcolor{white}{0.9384}} & \cellcolor[RGB]{34,176,37}\textbf{0.7616} & \cellcolor[RGB]{67,188,69}1.1311 & \cellcolor[RGB]{93,197,95}0.8459 & \cellcolor[RGB]{81,192,83}1.0253 & \cellcolor[RGB]{69,188,71}0.7899 \\
\hline
TiDE & \cellcolor[RGB]{122,207,124}1.2380 & \cellcolor[RGB]{207,238,208}0.9234 & \cellcolor[RGB]{126,209,128}1.1437 & \cellcolor[RGB]{208,238,208}0.8874 & \cellcolor[RGB]{207,237,207}1.3653 & \cellcolor[RGB]{255,242,242}0.9529 & \cellcolor[RGB]{255,152,152}1.2898 & \cellcolor[RGB]{255,85,85}0.9173 \\
\hline
SparseTSF & \cellcolor[RGB]{131,210,133}1.2592 & \cellcolor[RGB]{255,250,250}0.9588 & \cellcolor[RGB]{106,201,108}1.1007 & \cellcolor[RGB]{206,237,206}0.8858 & \cellcolor[RGB]{156,219,157}1.2797 & \cellcolor[RGB]{214,240,215}0.9209 & \cellcolor[RGB]{149,217,150}1.0938 & \cellcolor[RGB]{175,226,176}0.8305 \\
\hline
PatchTST & \cellcolor[RGB]{134,211,136}1.2665 & \cellcolor[RGB]{233,247,233}0.9413 & \cellcolor[RGB]{115,205,117}1.1207 & \cellcolor[RGB]{214,240,214}0.8918 & \cellcolor[RGB]{180,228,181}1.3207 & \cellcolor[RGB]{238,249,239}0.9361 & \cellcolor[RGB]{227,245,227}1.1721 & \cellcolor[RGB]{255,230,230}0.8688 \\
\hline
VCFormer & \cellcolor[RGB]{174,226,175}1.3602 & \cellcolor[RGB]{255,200,200}0.9892 & \cellcolor[RGB]{122,207,124}1.1354 & \cellcolor[RGB]{204,237,205}0.8850 & \cellcolor[RGB]{159,220,160}1.2851 & \cellcolor[RGB]{201,235,202}0.9128 & \cellcolor[RGB]{136,212,137}1.0805 & \cellcolor[RGB]{146,216,148}0.8194 \\
\hline
Autoformer & \cellcolor[RGB]{173,225,174}1.3595 & \cellcolor[RGB]{255,205,205}0.9860 & \cellcolor[RGB]{102,200,104}1.0929 & \cellcolor[RGB]{180,228,181}0.8672 & \cellcolor[RGB]{122,207,123}1.2229 & \cellcolor[RGB]{162,222,164}0.8888 & \cellcolor[RGB]{90,196,92}1.0344 & \cellcolor[RGB]{94,197,96}0.7995 \\
\hline
TimeMixer & \cellcolor[RGB]{34,176,37}\textbf{1.0287} & \cellcolor[RGB]{37,177,40}\textbf{0.8066} & \cellcolor[RGB]{145,216,147}1.1855 & \cellcolor[RGB]{176,226,177}0.8642 & \cellcolor[RGB]{255,246,246}1.4576 & \cellcolor[RGB]{249,253,249}0.9427 & \cellcolor[RGB]{224,244,224}1.1691 & \cellcolor[RGB]{197,234,198}0.8389 \\
\hline
WaveNet & \cellcolor[RGB]{255,0,0}2.0825 & \cellcolor[RGB]{255,0,0}1.1099 & \cellcolor[RGB]{255,112,112}1.6890 & \cellcolor[RGB]{255,125,125}1.0031 & \cellcolor[RGB]{255,0,0}1.8205 & \cellcolor[RGB]{255,0,0}1.0853 & \cellcolor[RGB]{255,61,61}1.3701 & \cellcolor[RGB]{255,67,67}0.9236 \\
\hline
FEDformer & \cellcolor[RGB]{210,238,210}1.4455 & \cellcolor[RGB]{255,179,179}1.0015 & \cellcolor[RGB]{255,0,0}1.9010 & \cellcolor[RGB]{255,0,0}1.0830 & \cellcolor[RGB]{255,199,199}1.5275 & \cellcolor[RGB]{255,195,195}0.9788 & \cellcolor[RGB]{255,0,0}1.4242 & \cellcolor[RGB]{255,0,0}0.9461 \\
\hline
iTransformer & \cellcolor[RGB]{176,226,177}1.3658 & \cellcolor[RGB]{255,202,202}0.9878 & \cellcolor[RGB]{110,203,112}1.1093 & \cellcolor[RGB]{201,235,201}0.8823 & \cellcolor[RGB]{155,219,156}1.2782 & \cellcolor[RGB]{195,233,196}0.9093 & \cellcolor[RGB]{146,216,148}1.0912 & \cellcolor[RGB]{169,224,170}0.8282 \\
\hline
TimeFlex & \cellcolor[RGB]{36,176,39}1.0344 & \cellcolor[RGB]{74,190,76}0.8321 & \cellcolor[RGB]{35,176,38}0.9489 & \cellcolor[RGB]{31,175,34}\textbf{\textcolor{white}{0.7594}} & \cellcolor[RGB]{88,195,91}1.1670 & \cellcolor[RGB]{155,219,156}0.8840 & \cellcolor[RGB]{169,224,170}1.1137 & \cellcolor[RGB]{163,222,165}0.8260 \\
\hline
\end{tabular}
}
\caption{Full results (MSE and MAE) of different models trained on the dataset with rational quadratic kernel.} \label{tab:rational}
\end{table}

%% file: tables/sampled/combined_kernel_table.tex
\begin{table}[htbp]
\centering
\resizebox{\columnwidth}{!}{
\begin{tabular}{|l|rr|rr|rr|rr|}
\hline
$\empty$ & \multicolumn{2}{c|}{\textbf{96 steps}} & \multicolumn{2}{c|}{\textbf{196 steps}} & \multicolumn{2}{c|}{\textbf{336 steps}} & \multicolumn{2}{c|}{\textbf{720 steps}} \\
\hline
\textbf{Model} & \textbf{MSE} & \textbf{MAE} & \textbf{MSE} & \textbf{MAE} & \textbf{MSE} & \textbf{MAE} & \textbf{MSE} & \textbf{MAE} \\
\hline
CycleNet & \cellcolor[RGB]{31,175,34}\textbf{\textcolor{white}{0.8486}} & \cellcolor[RGB]{31,175,34}\textbf{\textcolor{white}{0.6893}} & \cellcolor[RGB]{42,179,45}1.3047 & \cellcolor[RGB]{35,176,38}0.8917 & \cellcolor[RGB]{47,180,50}1.1052 & \cellcolor[RGB]{54,183,57}0.7981 & \cellcolor[RGB]{47,180,50}0.9460 & \cellcolor[RGB]{48,181,50}0.7420 \\
\hline
WaveMask & \cellcolor[RGB]{34,176,37}0.8617 & \cellcolor[RGB]{43,179,46}0.7100 & \cellcolor[RGB]{39,178,42}1.2943 & \cellcolor[RGB]{31,175,34}\textbf{\textcolor{white}{0.8862}} & \cellcolor[RGB]{32,175,35}\textbf{1.0607} & \cellcolor[RGB]{35,176,38}\textbf{0.7791} & \cellcolor[RGB]{31,175,34}\textbf{\textcolor{white}{0.9110}} & \cellcolor[RGB]{31,175,34}\textbf{\textcolor{white}{0.7281}} \\
\hline
DLinear & \cellcolor[RGB]{34,176,37}\textbf{0.8612} & \cellcolor[RGB]{42,179,45}\textbf{0.7081} & \cellcolor[RGB]{38,177,41}\textbf{1.2894} & \cellcolor[RGB]{31,175,34}\textbf{0.8872} & \cellcolor[RGB]{31,175,34}\textbf{\textcolor{white}{1.0547}} & \cellcolor[RGB]{31,175,34}\textbf{\textcolor{white}{0.7747}} & \cellcolor[RGB]{33,175,36}\textbf{0.9159} & \cellcolor[RGB]{34,176,37}\textbf{0.7313} \\
\hline
RLinear & \cellcolor[RGB]{180,228,181}1.4634 & \cellcolor[RGB]{171,225,172}0.9191 & \cellcolor[RGB]{255,248,248}2.1335 & \cellcolor[RGB]{255,247,247}1.1444 & \cellcolor[RGB]{236,248,236}1.6835 & \cellcolor[RGB]{255,233,233}1.0124 & \cellcolor[RGB]{255,222,222}1.4439 & \cellcolor[RGB]{255,217,217}0.9369 \\
\hline
NLinear & \cellcolor[RGB]{176,226,177}1.4467 & \cellcolor[RGB]{162,221,163}0.9047 & \cellcolor[RGB]{255,236,236}2.1755 & \cellcolor[RGB]{255,239,239}1.1519 & \cellcolor[RGB]{245,251,245}1.7118 & \cellcolor[RGB]{255,229,229}1.0154 & \cellcolor[RGB]{255,206,206}1.4726 & \cellcolor[RGB]{255,206,206}0.9450 \\
\hline
Crossformer & \cellcolor[RGB]{67,188,70}1.0004 & \cellcolor[RGB]{113,204,115}0.8239 & \cellcolor[RGB]{51,182,54}1.3390 & \cellcolor[RGB]{85,194,87}0.9467 & \cellcolor[RGB]{38,177,41}1.0781 & \cellcolor[RGB]{52,182,55}0.7956 & \cellcolor[RGB]{47,181,50}0.9467 & \cellcolor[RGB]{70,189,72}0.7602 \\
\hline
TiDE & \cellcolor[RGB]{164,222,166}1.4004 & \cellcolor[RGB]{125,208,127}0.8450 & \cellcolor[RGB]{255,236,236}2.1730 & \cellcolor[RGB]{246,251,246}1.1272 & \cellcolor[RGB]{254,254,254}1.7391 & \cellcolor[RGB]{255,239,239}1.0070 & \cellcolor[RGB]{255,215,215}1.4554 & \cellcolor[RGB]{255,232,232}0.9268 \\
\hline
SparseTSF & \cellcolor[RGB]{255,219,219}1.9002 & \cellcolor[RGB]{255,234,234}1.0865 & \cellcolor[RGB]{255,176,176}2.3754 & \cellcolor[RGB]{255,161,161}1.2296 & \cellcolor[RGB]{255,200,200}1.8874 & \cellcolor[RGB]{255,147,147}1.0862 & \cellcolor[RGB]{255,154,154}1.5689 & \cellcolor[RGB]{255,163,163}0.9756 \\
\hline
PatchTST & \cellcolor[RGB]{165,223,166}1.4027 & \cellcolor[RGB]{159,221,161}0.9008 & \cellcolor[RGB]{255,195,195}2.3127 & \cellcolor[RGB]{255,196,196}1.1946 & \cellcolor[RGB]{247,252,247}1.7173 & \cellcolor[RGB]{255,228,228}1.0160 & \cellcolor[RGB]{255,195,195}1.4939 & \cellcolor[RGB]{255,207,207}0.9443 \\
\hline
VCFormer & \cellcolor[RGB]{255,197,197}1.9798 & \cellcolor[RGB]{255,207,207}1.1246 & \cellcolor[RGB]{255,176,176}2.3753 & \cellcolor[RGB]{255,144,144}1.2454 & \cellcolor[RGB]{255,207,207}1.8672 & \cellcolor[RGB]{255,151,151}1.0825 & \cellcolor[RGB]{255,180,180}1.5208 & \cellcolor[RGB]{255,183,183}0.9614 \\
\hline
Autoformer & \cellcolor[RGB]{255,157,157}2.1254 & \cellcolor[RGB]{255,191,191}1.1482 & \cellcolor[RGB]{255,127,127}2.5396 & \cellcolor[RGB]{255,123,123}1.2664 & \cellcolor[RGB]{255,182,182}1.9357 & \cellcolor[RGB]{255,138,138}1.0938 & \cellcolor[RGB]{255,160,160}1.5572 & \cellcolor[RGB]{255,174,174}0.9677 \\
\hline
TimeMixer & \cellcolor[RGB]{255,198,198}1.9759 & \cellcolor[RGB]{255,205,205}1.1281 & \cellcolor[RGB]{255,0,0}2.9641 & \cellcolor[RGB]{255,23,23}1.3651 & \cellcolor[RGB]{255,48,48}2.2964 & \cellcolor[RGB]{255,0,0}1.2125 & \cellcolor[RGB]{255,0,0}1.8548 & \cellcolor[RGB]{255,0,0}1.0927 \\
\hline
WaveNet & \cellcolor[RGB]{191,232,192}1.5105 & \cellcolor[RGB]{224,244,225}1.0068 & \cellcolor[RGB]{84,194,86}1.4637 & \cellcolor[RGB]{96,198,98}0.9596 & \cellcolor[RGB]{139,213,141}1.3876 & \cellcolor[RGB]{187,230,188}0.9274 & \cellcolor[RGB]{137,212,138}1.1347 & \cellcolor[RGB]{130,210,131}0.8087 \\
\hline
FEDformer & \cellcolor[RGB]{255,0,0}2.6944 & \cellcolor[RGB]{255,0,0}1.4240 & \cellcolor[RGB]{255,60,60}2.7626 & \cellcolor[RGB]{255,0,0}1.3880 & \cellcolor[RGB]{255,0,0}2.4268 & \cellcolor[RGB]{255,67,67}1.1547 & \cellcolor[RGB]{255,104,104}1.6621 & \cellcolor[RGB]{255,87,87}1.0300 \\
\hline
iTransformer & \cellcolor[RGB]{255,236,236}1.8398 & \cellcolor[RGB]{255,243,243}1.0731 & \cellcolor[RGB]{255,160,160}2.4274 & \cellcolor[RGB]{255,143,143}1.2471 & \cellcolor[RGB]{255,193,193}1.9074 & \cellcolor[RGB]{255,128,128}1.1020 & \cellcolor[RGB]{255,206,206}1.4719 & \cellcolor[RGB]{255,209,209}0.9429 \\
\hline
TimeFlex & \cellcolor[RGB]{38,177,40}0.8778 & \cellcolor[RGB]{69,188,72}0.7527 & \cellcolor[RGB]{31,175,34}\textbf{\textcolor{white}{1.2611}} & \cellcolor[RGB]{40,178,42}0.8964 & \cellcolor[RGB]{111,203,113}1.3011 & \cellcolor[RGB]{120,206,121}0.8619 & \cellcolor[RGB]{62,186,65}0.9775 & \cellcolor[RGB]{71,189,73}0.7611 \\
\hline
\end{tabular}
}
\caption{Full results (MSE and MAE) of different models trained on the dataset with combined kernel.} \label{tab:combined}
\end{table}

%% file: tables/sampled/se_kernel_table.tex
\begin{table}[htbp]
\centering
\resizebox{\columnwidth}{!}{
\begin{tabular}{|l|rr|rr|rr|rr|}
\hline
$\empty$ & \multicolumn{2}{c|}{\textbf{96 steps}} & \multicolumn{2}{c|}{\textbf{196 steps}} & \multicolumn{2}{c|}{\textbf{336 steps}} & \multicolumn{2}{c|}{\textbf{720 steps}} \\
\hline
\textbf{model} & \textbf{MSE} & \textbf{MAE} & \textbf{MSE} & \textbf{MAE} & \textbf{MSE} & \textbf{MAE} & \textbf{MSE} & \textbf{MAE} \\
\hline
CycleNet & \cellcolor[RGB]{31,175,34}\textbf{\textcolor{white}{0.4734}} & \cellcolor[RGB]{35,176,38}\textbf{0.5421} & \cellcolor[RGB]{112,203,114}0.7156 & \cellcolor[RGB]{142,214,143}0.6767 & \cellcolor[RGB]{39,178,42}0.7847 & \cellcolor[RGB]{46,180,49}0.7203 & \cellcolor[RGB]{34,176,37}0.8707 & \cellcolor[RGB]{38,177,41}0.7561 \\
\hline
WaveMask & \cellcolor[RGB]{34,176,37}0.4860 & \cellcolor[RGB]{40,178,43}0.5485 & \cellcolor[RGB]{90,196,92}0.6735 & \cellcolor[RGB]{128,209,129}0.6641 & \cellcolor[RGB]{31,175,34}\textbf{\textcolor{white}{0.7641}} & \cellcolor[RGB]{37,177,40}\textbf{0.7118} & \cellcolor[RGB]{31,175,34}\textbf{0.8645} & \cellcolor[RGB]{32,175,35}\textbf{0.7522} \\
\hline
DLinear & \cellcolor[RGB]{35,176,38}0.4883 & \cellcolor[RGB]{42,179,45}0.5511 & \cellcolor[RGB]{94,197,97}0.6825 & \cellcolor[RGB]{128,209,130}0.6644 & \cellcolor[RGB]{36,176,39}\textbf{0.7774} & \cellcolor[RGB]{42,179,45}0.7168 & \cellcolor[RGB]{31,175,34}\textbf{\textcolor{white}{0.8633}} & \cellcolor[RGB]{31,175,34}\textbf{\textcolor{white}{0.7508}} \\
\hline
RLinear & \cellcolor[RGB]{71,189,73}0.6119 & \cellcolor[RGB]{111,203,113}0.6362 & \cellcolor[RGB]{54,183,57}0.6048 & \cellcolor[RGB]{85,194,87}0.6254 & \cellcolor[RGB]{70,189,73}0.8597 & \cellcolor[RGB]{72,189,74}0.7443 & \cellcolor[RGB]{144,215,145}1.0760 & \cellcolor[RGB]{136,212,138}0.8285 \\
\hline
NLinear & \cellcolor[RGB]{45,180,48}0.5238 & \cellcolor[RGB]{40,178,43}0.5483 & \cellcolor[RGB]{31,175,34}\textbf{\textcolor{white}{0.5597}} & \cellcolor[RGB]{34,176,37}\textbf{0.5800} & \cellcolor[RGB]{54,183,57}0.8210 & \cellcolor[RGB]{42,178,44}0.7162 & \cellcolor[RGB]{135,212,136}1.0587 & \cellcolor[RGB]{122,207,123}0.8178 \\
\hline
Crossformer & \cellcolor[RGB]{43,179,46}0.5171 & \cellcolor[RGB]{74,190,77}0.5907 & \cellcolor[RGB]{105,201,107}0.7034 & \cellcolor[RGB]{161,221,162}0.6941 & \cellcolor[RGB]{53,183,56}0.8190 & \cellcolor[RGB]{83,193,85}0.7543 & \cellcolor[RGB]{41,178,44}0.8828 & \cellcolor[RGB]{57,184,60}0.7706 \\
\hline
TiDE & \cellcolor[RGB]{47,180,50}0.5295 & \cellcolor[RGB]{55,183,57}0.5664 & \cellcolor[RGB]{33,176,36}\textbf{0.5655} & \cellcolor[RGB]{31,175,34}\textbf{\textcolor{white}{0.5765}} & \cellcolor[RGB]{54,183,57}0.8213 & \cellcolor[RGB]{40,178,43}0.7151 & \cellcolor[RGB]{142,214,143}1.0719 & \cellcolor[RGB]{130,210,131}0.8236 \\
\hline
SparseTSF & \cellcolor[RGB]{122,207,124}0.7887 & \cellcolor[RGB]{188,231,189}0.7315 & \cellcolor[RGB]{107,202,109}0.7066 & \cellcolor[RGB]{115,205,117}0.6524 & \cellcolor[RGB]{112,204,114}0.9613 & \cellcolor[RGB]{105,201,107}0.7752 & \cellcolor[RGB]{197,234,197}1.1753 & \cellcolor[RGB]{187,230,188}0.8657 \\
\hline
PatchTST & \cellcolor[RGB]{70,189,72}0.6085 & \cellcolor[RGB]{119,206,121}0.6463 & \cellcolor[RGB]{119,206,121}0.7295 & \cellcolor[RGB]{128,209,130}0.6647 & \cellcolor[RGB]{64,187,67}0.8459 & \cellcolor[RGB]{56,184,59}0.7300 & \cellcolor[RGB]{159,220,160}1.1040 & \cellcolor[RGB]{153,218,155}0.8413 \\
\hline
VCFormer & \cellcolor[RGB]{129,210,131}0.8129 & \cellcolor[RGB]{206,237,207}0.7542 & \cellcolor[RGB]{126,209,127}0.7426 & \cellcolor[RGB]{136,212,138}0.6719 & \cellcolor[RGB]{125,208,127}0.9921 & \cellcolor[RGB]{124,208,125}0.7921 & \cellcolor[RGB]{212,239,212}1.2032 & \cellcolor[RGB]{205,237,206}0.8795 \\
\hline
Autoformer & \cellcolor[RGB]{173,225,174}0.9640 & \cellcolor[RGB]{247,252,247}0.8050 & \cellcolor[RGB]{142,214,144}0.7745 & \cellcolor[RGB]{152,218,153}0.6857 & \cellcolor[RGB]{130,210,132}1.0043 & \cellcolor[RGB]{130,210,132}0.7979 & \cellcolor[RGB]{226,244,226}1.2305 & \cellcolor[RGB]{222,243,222}0.8916 \\
\hline
TimeMixer & \cellcolor[RGB]{93,197,95}0.6877 & \cellcolor[RGB]{114,204,115}0.6394 & \cellcolor[RGB]{95,197,97}0.6830 & \cellcolor[RGB]{124,208,126}0.6606 & \cellcolor[RGB]{37,177,40}0.7805 & \cellcolor[RGB]{31,175,34}\textbf{\textcolor{white}{0.7060}} & \cellcolor[RGB]{143,215,144}1.0740 & \cellcolor[RGB]{137,213,139}0.8291 \\
\hline
WaveNet & \cellcolor[RGB]{202,236,203}1.0630 & \cellcolor[RGB]{255,179,179}0.8962 & \cellcolor[RGB]{255,0,0}1.4202 & \cellcolor[RGB]{255,0,0}0.9799 & \cellcolor[RGB]{166,223,167}1.0904 & \cellcolor[RGB]{173,225,174}0.8378 & \cellcolor[RGB]{137,213,139}1.0637 & \cellcolor[RGB]{153,218,155}0.8413 \\
\hline
FEDformer & \cellcolor[RGB]{255,0,0}2.0153 & \cellcolor[RGB]{255,0,0}1.0922 & \cellcolor[RGB]{255,91,91}1.2655 & \cellcolor[RGB]{255,115,115}0.8884 & \cellcolor[RGB]{255,0,0}1.8465 & \cellcolor[RGB]{255,0,0}1.1202 & \cellcolor[RGB]{255,0,0}1.7045 & \cellcolor[RGB]{255,0,0}1.0803 \\
\hline
iTransformer & \cellcolor[RGB]{135,212,137}0.8333 & \cellcolor[RGB]{211,239,212}0.7604 & \cellcolor[RGB]{145,215,146}0.7790 & \cellcolor[RGB]{142,214,143}0.6766 & \cellcolor[RGB]{111,203,113}0.9584 & \cellcolor[RGB]{110,203,112}0.7794 & \cellcolor[RGB]{223,243,224}1.2256 & \cellcolor[RGB]{212,239,213}0.8846 \\
\hline
TimeFlex & \cellcolor[RGB]{32,175,35}\textbf{0.4783} & \cellcolor[RGB]{31,175,34}\textbf{\textcolor{white}{0.5365}} & \cellcolor[RGB]{110,203,112}0.7122 & \cellcolor[RGB]{166,223,167}0.6985 & \cellcolor[RGB]{85,194,87}0.8963 & \cellcolor[RGB]{89,195,91}0.7600 & \cellcolor[RGB]{68,188,70}0.9334 & \cellcolor[RGB]{57,184,60}0.7702 \\
\hline
\end{tabular}
}
\caption{Full results (MSE and MAE) of different models trained on the dataset with se-kernel.} \label{tab:se}
\end{table}

%% file: tables/bench/weather_table.tex
\begin{table}[htbp]
\centering
\resizebox{\columnwidth}{!}{
\begin{tabular}{|l|rr|rr|rr|rr|}
\hline
$\empty$ & \multicolumn{2}{c|}{\textbf{96 steps}} & \multicolumn{2}{c|}{\textbf{196 steps}} & \multicolumn{2}{c|}{\textbf{336 steps}} & \multicolumn{2}{c|}{\textbf{720 steps}} \\
\hline
\textbf{Model} & \textbf{MSE} & \textbf{MAE} & \textbf{MSE.1} & \textbf{MAE.1} & \textbf{MSE.2} & \textbf{MAE.2} & \textbf{MSE.3} & \textbf{MAE.3} \\
\hline
CycleNet & \cellcolor[RGB]{226,244,226}0.1404 & \cellcolor[RGB]{255,205,205}0.2557 & \cellcolor[RGB]{125,208,126}0.1189 & \cellcolor[RGB]{223,243,223}0.2465 & \cellcolor[RGB]{225,244,225}0.1836 & \cellcolor[RGB]{255,119,119}0.3036 & \cellcolor[RGB]{255,70,70}0.2960 & \cellcolor[RGB]{255,46,46}0.4005 \\
\hline
WaveMask & \cellcolor[RGB]{113,204,115}0.0944 & \cellcolor[RGB]{199,235,200}0.2217 & \cellcolor[RGB]{92,197,95}0.1012 & \cellcolor[RGB]{174,226,175}0.2240 & \cellcolor[RGB]{138,213,139}0.1469 & \cellcolor[RGB]{255,251,251}0.2585 & \cellcolor[RGB]{188,231,189}0.2104 & \cellcolor[RGB]{255,234,234}0.3271 \\
\hline
DLinear & \cellcolor[RGB]{155,219,156}0.1116 & \cellcolor[RGB]{255,233,233}0.2473 & \cellcolor[RGB]{119,206,121}0.1160 & \cellcolor[RGB]{227,245,228}0.2487 & \cellcolor[RGB]{183,229,184}0.1658 & \cellcolor[RGB]{255,170,170}0.2863 & \cellcolor[RGB]{234,247,234}0.2277 & \cellcolor[RGB]{255,192,192}0.3434 \\
\hline
RLinear & \cellcolor[RGB]{66,187,68}0.0753 & \cellcolor[RGB]{100,199,102}0.1877 & \cellcolor[RGB]{40,178,43}0.0724 & \cellcolor[RGB]{50,181,52}0.1669 & \cellcolor[RGB]{37,177,39}0.1042 & \cellcolor[RGB]{58,184,61}0.1807 & \cellcolor[RGB]{46,180,49}\textbf{0.1571} & \cellcolor[RGB]{48,181,51}\textbf{0.2272} \\
\hline
NLinear & \cellcolor[RGB]{64,186,66}0.0745 & \cellcolor[RGB]{86,194,88}0.1829 & \cellcolor[RGB]{40,178,43}0.0726 & \cellcolor[RGB]{35,176,38}0.1602 & \cellcolor[RGB]{42,179,45}0.1064 & \cellcolor[RGB]{50,182,53}0.1777 & \cellcolor[RGB]{57,184,59}0.1611 & \cellcolor[RGB]{56,183,58}0.2305 \\
\hline
Crossformer & \cellcolor[RGB]{138,213,140}0.1047 & \cellcolor[RGB]{107,202,109}0.1903 & \cellcolor[RGB]{235,248,236}0.1800 & \cellcolor[RGB]{255,236,236}0.2685 & \cellcolor[RGB]{255,32,32}0.2787 & \cellcolor[RGB]{255,0,0}0.3446 & \cellcolor[RGB]{255,84,84}0.2915 & \cellcolor[RGB]{255,178,178}0.3490 \\
\hline
TiDE & \cellcolor[RGB]{31,175,34}\textbf{\textcolor{white}{0.0610}} & \cellcolor[RGB]{31,175,34}\textbf{\textcolor{white}{0.1640}} & \cellcolor[RGB]{31,175,34}\textbf{0.0672} & \cellcolor[RGB]{31,175,34}\textbf{0.1584} & \cellcolor[RGB]{33,175,36}0.1025 & \cellcolor[RGB]{47,180,50}\textbf{0.1764} & \cellcolor[RGB]{46,180,49}0.1572 & \cellcolor[RGB]{57,184,59}0.2310 \\
\hline
SparseTSF & \cellcolor[RGB]{44,179,46}0.0664 & \cellcolor[RGB]{61,185,63}0.1743 & \cellcolor[RGB]{40,178,43}0.0726 & \cellcolor[RGB]{53,183,56}0.1686 & \cellcolor[RGB]{46,180,48}0.1080 & \cellcolor[RGB]{72,189,74}0.1859 & \cellcolor[RGB]{61,186,64}0.1629 & \cellcolor[RGB]{76,191,78}0.2396 \\
\hline
PatchTST & \cellcolor[RGB]{76,191,78}0.0796 & \cellcolor[RGB]{95,198,97}0.1862 & \cellcolor[RGB]{36,176,39}0.0700 & \cellcolor[RGB]{31,175,34}\textbf{\textcolor{white}{0.1581}} & \cellcolor[RGB]{32,175,35}\textbf{0.1021} & \cellcolor[RGB]{31,175,34}\textbf{\textcolor{white}{0.1700}} & \cellcolor[RGB]{60,185,63}0.1624 & \cellcolor[RGB]{62,186,64}0.2333 \\
\hline
Autoformer & \cellcolor[RGB]{210,239,211}0.1340 & \cellcolor[RGB]{255,138,138}0.2758 & \cellcolor[RGB]{113,204,115}0.1124 & \cellcolor[RGB]{230,246,231}0.2501 & \cellcolor[RGB]{91,196,93}0.1270 & \cellcolor[RGB]{224,244,225}0.2455 & \cellcolor[RGB]{255,46,46}0.3039 & \cellcolor[RGB]{255,75,75}0.3894 \\
\hline
TimeMixer & \cellcolor[RGB]{42,179,45}\textbf{0.0659} & \cellcolor[RGB]{54,183,57}\textbf{0.1721} & \cellcolor[RGB]{56,184,59}0.0812 & \cellcolor[RGB]{81,193,84}0.1815 & \cellcolor[RGB]{106,201,108}0.1334 & \cellcolor[RGB]{173,225,174}0.2255 & \cellcolor[RGB]{59,185,62}0.1620 & \cellcolor[RGB]{75,190,77}0.2390 \\
\hline
WaveNet & \cellcolor[RGB]{255,0,0}0.2430 & \cellcolor[RGB]{255,0,0}0.3176 & \cellcolor[RGB]{255,0,0}0.3141 & \cellcolor[RGB]{255,0,0}0.3641 & \cellcolor[RGB]{255,0,0}0.2906 & \cellcolor[RGB]{255,13,13}0.3399 & \cellcolor[RGB]{255,216,216}0.2481 & \cellcolor[RGB]{231,246,232}0.3088 \\
\hline
FEDformer & \cellcolor[RGB]{194,233,195}0.1276 & \cellcolor[RGB]{255,199,199}0.2574 & \cellcolor[RGB]{104,201,106}0.1078 & \cellcolor[RGB]{195,233,196}0.2337 & \cellcolor[RGB]{112,204,114}0.1361 & \cellcolor[RGB]{255,246,246}0.2602 & \cellcolor[RGB]{255,0,0}0.3193 & \cellcolor[RGB]{255,0,0}0.4188 \\
\hline
iTransformer & \cellcolor[RGB]{86,194,89}0.0838 & \cellcolor[RGB]{101,200,103}0.1882 & \cellcolor[RGB]{31,175,34}\textbf{\textcolor{white}{0.0671}} & \cellcolor[RGB]{38,177,41}0.1616 & \cellcolor[RGB]{75,190,77}0.1202 & \cellcolor[RGB]{107,202,109}0.1998 & \cellcolor[RGB]{131,210,133}0.1890 & \cellcolor[RGB]{144,215,146}0.2700 \\
\hline
TimeFlex & \cellcolor[RGB]{48,181,51}0.0683 & \cellcolor[RGB]{90,196,92}0.1843 & \cellcolor[RGB]{37,177,40}0.0709 & \cellcolor[RGB]{38,177,41}0.1614 & \cellcolor[RGB]{31,175,34}\textbf{\textcolor{white}{0.1016}} & \cellcolor[RGB]{58,184,61}0.1808 & \cellcolor[RGB]{31,175,34}\textbf{\textcolor{white}{0.1513}} & \cellcolor[RGB]{31,175,34}\textbf{\textcolor{white}{0.2193}} \\
\hline
\end{tabular}
}
\caption{Full results (MSE and MAE) of different models trained on the Weather dataset.} \label{tab:weather}
\end{table}

%% file: tables/bench/Etth1_table.tex
\begin{table}[htbp]
\centering
\resizebox{\columnwidth}{!}{
\begin{tabular}{|l|rr|rr|rr|rr|}
\hline
$\empty$ & \multicolumn{2}{c|}{\textbf{92 steps}} & \multicolumn{2}{c|}{\textbf{196 steps}} & \multicolumn{2}{c|}{\textbf{336 steps}} & \multicolumn{2}{c|}{\textbf{720 steps}} \\
\hline
\textbf{Model} & \textbf{MSE} & \textbf{MAE} & \textbf{MSE.1} & \textbf{MAE.1} & \textbf{MSE.2} & \textbf{MAE.2} & \textbf{MSE.3} & \textbf{MAE.3} \\
\hline
CycleNet & \cellcolor[RGB]{43,179,46}0.2525 & \cellcolor[RGB]{57,184,60}0.3333 & \cellcolor[RGB]{41,178,43}0.3205 & \cellcolor[RGB]{41,178,44}0.3817 & \cellcolor[RGB]{31,175,34}\textbf{\textcolor{white}{0.2903}} & \cellcolor[RGB]{31,175,34}\textbf{\textcolor{white}{0.3777}} & \cellcolor[RGB]{31,175,34}\textbf{\textcolor{white}{0.4928}} & \cellcolor[RGB]{31,175,34}\textbf{\textcolor{white}{0.5118}} \\
\hline
WaveMask & \cellcolor[RGB]{35,176,38}\textbf{0.2403} & \cellcolor[RGB]{56,184,59}\textbf{0.3323} & \cellcolor[RGB]{31,175,34}\textbf{\textcolor{white}{0.3078}} & \cellcolor[RGB]{31,175,34}\textbf{\textcolor{white}{0.3734}} & \cellcolor[RGB]{59,185,62}0.3256 & \cellcolor[RGB]{66,187,69}0.4047 & \cellcolor[RGB]{55,183,58}\textbf{0.5300} & \cellcolor[RGB]{71,189,73}0.5416 \\
\hline
DLinear & \cellcolor[RGB]{48,181,51}0.2589 & \cellcolor[RGB]{64,186,67}0.3395 & \cellcolor[RGB]{46,180,49}0.3271 & \cellcolor[RGB]{35,176,38}\textbf{0.3769} & \cellcolor[RGB]{66,187,68}0.3333 & \cellcolor[RGB]{62,186,65}0.4018 & \cellcolor[RGB]{61,185,64}0.5390 & \cellcolor[RGB]{66,187,69}0.5385 \\
\hline
RLinear & \cellcolor[RGB]{52,182,55}0.2653 & \cellcolor[RGB]{61,185,64}0.3368 & \cellcolor[RGB]{88,195,90}0.3801 & \cellcolor[RGB]{72,189,74}0.4055 & \cellcolor[RGB]{78,191,80}0.3481 & \cellcolor[RGB]{75,190,77}0.4113 & \cellcolor[RGB]{88,195,90}0.5794 & \cellcolor[RGB]{81,192,83}0.5493 \\
\hline
NLinear & \cellcolor[RGB]{67,187,69}0.2858 & \cellcolor[RGB]{69,188,72}0.3443 & \cellcolor[RGB]{106,202,108}0.4033 & \cellcolor[RGB]{87,195,89}0.4171 & \cellcolor[RGB]{103,200,105}0.3789 & \cellcolor[RGB]{94,197,96}0.4257 & \cellcolor[RGB]{97,198,100}0.5939 & \cellcolor[RGB]{99,199,101}0.5630 \\
\hline
Crossformer & \cellcolor[RGB]{98,199,100}0.3303 & \cellcolor[RGB]{121,207,122}0.3895 & \cellcolor[RGB]{35,176,38}\textbf{0.3131} & \cellcolor[RGB]{73,190,75}0.4063 & \cellcolor[RGB]{112,204,114}0.3901 & \cellcolor[RGB]{144,215,145}0.4631 & \cellcolor[RGB]{165,223,166}0.6961 & \cellcolor[RGB]{168,223,169}0.6140 \\
\hline
TiDE & \cellcolor[RGB]{31,175,34}\textbf{\textcolor{white}{0.2344}} & \cellcolor[RGB]{31,175,34}\textbf{\textcolor{white}{0.3098}} & \cellcolor[RGB]{55,183,57}0.3384 & \cellcolor[RGB]{42,179,45}0.3821 & \cellcolor[RGB]{44,179,46}0.3065 & \cellcolor[RGB]{40,178,43}\textbf{0.3849} & \cellcolor[RGB]{76,191,78}0.5609 & \cellcolor[RGB]{67,188,70}0.5390 \\
\hline
SparseTSF & \cellcolor[RGB]{39,177,42}0.2460 & \cellcolor[RGB]{61,185,64}0.3367 & \cellcolor[RGB]{46,180,49}0.3278 & \cellcolor[RGB]{56,184,59}0.3931 & \cellcolor[RGB]{41,178,44}\textbf{0.3029} & \cellcolor[RGB]{52,182,55}0.3938 & \cellcolor[RGB]{64,187,67}0.5436 & \cellcolor[RGB]{61,185,63}\textbf{0.5342} \\
\hline
PatchTST & \cellcolor[RGB]{43,179,46}0.2520 & \cellcolor[RGB]{90,196,92}0.3625 & \cellcolor[RGB]{80,192,82}0.3704 & \cellcolor[RGB]{126,209,128}0.4477 & \cellcolor[RGB]{94,197,96}0.3678 & \cellcolor[RGB]{138,213,139}0.4586 & \cellcolor[RGB]{144,215,145}0.6636 & \cellcolor[RGB]{179,227,180}0.6224 \\
\hline
Autoformer & \cellcolor[RGB]{255,218,218}0.5962 & \cellcolor[RGB]{255,109,109}0.6209 & \cellcolor[RGB]{255,189,189}0.6630 & \cellcolor[RGB]{255,115,115}0.6427 & \cellcolor[RGB]{214,240,215}0.5160 & \cellcolor[RGB]{255,205,205}0.5797 & \cellcolor[RGB]{255,34,34}1.1233 & \cellcolor[RGB]{255,28,28}0.8269 \\
\hline
TimeMixer & \cellcolor[RGB]{62,186,64}0.2787 & \cellcolor[RGB]{100,199,102}0.3716 & \cellcolor[RGB]{100,199,102}0.3956 & \cellcolor[RGB]{135,212,137}0.4545 & \cellcolor[RGB]{104,201,106}0.3811 & \cellcolor[RGB]{146,216,148}0.4651 & \cellcolor[RGB]{255,51,51}1.1007 & \cellcolor[RGB]{255,63,63}0.8045 \\
\hline
WaveNet & \cellcolor[RGB]{255,237,237}0.5731 & \cellcolor[RGB]{255,254,254}0.5081 & \cellcolor[RGB]{255,207,207}0.6428 & \cellcolor[RGB]{255,205,205}0.5810 & \cellcolor[RGB]{255,74,74}0.7596 & \cellcolor[RGB]{255,152,152}0.6146 & \cellcolor[RGB]{245,251,245}0.8162 & \cellcolor[RGB]{215,240,216}0.6494 \\
\hline
FEDformer & \cellcolor[RGB]{125,208,126}0.3673 & \cellcolor[RGB]{205,237,206}0.4641 & \cellcolor[RGB]{180,228,181}0.4960 & \cellcolor[RGB]{255,249,249}0.5510 & \cellcolor[RGB]{255,0,0}0.8400 & \cellcolor[RGB]{255,0,0}0.7158 & \cellcolor[RGB]{247,252,247}0.8192 & \cellcolor[RGB]{255,218,218}0.7027 \\
\hline
iTransformer & \cellcolor[RGB]{92,197,95}0.3219 & \cellcolor[RGB]{175,226,176}0.4375 & \cellcolor[RGB]{63,186,65}0.3485 & \cellcolor[RGB]{125,208,127}0.4467 & \cellcolor[RGB]{146,216,148}0.4324 & \cellcolor[RGB]{219,242,220}0.5200 & \cellcolor[RGB]{142,214,144}0.6614 & \cellcolor[RGB]{166,223,167}0.6128 \\
\hline
TimeFlex & \cellcolor[RGB]{46,180,48}0.2557 & \cellcolor[RGB]{86,194,88}0.3589 & \cellcolor[RGB]{41,178,44}0.3213 & \cellcolor[RGB]{113,204,115}0.4374 & \cellcolor[RGB]{91,196,93}0.3644 & \cellcolor[RGB]{155,219,156}0.4716 & \cellcolor[RGB]{92,196,94}0.5854 & \cellcolor[RGB]{138,213,139}0.5916 \\
\hline
\end{tabular}
}
\caption{Full results (MSE and MAE) of different models trained on the ETTh1 dataset.} \label{tab:ETTh1}
\end{table}

%% file: tables/bench/Ettm2_table.tex
\begin{table}[htbp]
\centering
\resizebox{\columnwidth}{!}{
\begin{tabular}{|l|rr|rr|rr|rr|}
\hline
$\empty$ & \multicolumn{2}{c|}{\textbf{96 steps}} & \multicolumn{2}{c|}{\textbf{196 steps}} & \multicolumn{2}{c|}{\textbf{336 steps}} & \multicolumn{2}{c|}{\textbf{720 steps}} \\
\hline
\textbf{Model} & \textbf{MSE} & \textbf{MAE} & \textbf{MSE.1} & \textbf{MAE.1} & \textbf{MSE.2} & \textbf{MAE.2} & \textbf{MSE.3} & \textbf{MAE.3} \\
\hline
CycleNet & \cellcolor[RGB]{86,194,89}0.2105 & \cellcolor[RGB]{91,196,93}0.3256 & \cellcolor[RGB]{177,227,178}0.3750 & \cellcolor[RGB]{205,237,205}0.4765 & \cellcolor[RGB]{135,212,136}0.4268 & \cellcolor[RGB]{196,234,197}0.5400 & \cellcolor[RGB]{186,230,186}0.5706 & \cellcolor[RGB]{255,248,248}0.6396 \\
\hline
WaveMask & \cellcolor[RGB]{31,175,34}\textbf{\textcolor{white}{0.1620}} & \cellcolor[RGB]{31,175,34}\textbf{\textcolor{white}{0.2827}} & \cellcolor[RGB]{66,187,69}\textbf{0.2685} & \cellcolor[RGB]{60,185,63}\textbf{0.3992} & \cellcolor[RGB]{31,175,34}\textbf{\textcolor{white}{0.3120}} & \cellcolor[RGB]{31,175,34}\textbf{\textcolor{white}{0.4472}} & \cellcolor[RGB]{86,194,88}\textbf{0.4281} & \cellcolor[RGB]{125,208,126}\textbf{0.5322} \\
\hline
DLinear & \cellcolor[RGB]{57,184,60}0.1853 & \cellcolor[RGB]{48,181,51}\textbf{0.2951} & \cellcolor[RGB]{82,193,84}0.2831 & \cellcolor[RGB]{71,189,73}0.4049 & \cellcolor[RGB]{49,181,52}\textbf{0.3326} & \cellcolor[RGB]{61,186,64}\textbf{0.4645} & \cellcolor[RGB]{95,198,97}0.4409 & \cellcolor[RGB]{136,212,137}0.5411 \\
\hline
RLinear & \cellcolor[RGB]{162,221,163}0.2758 & \cellcolor[RGB]{109,202,111}0.3382 & \cellcolor[RGB]{255,174,174}0.5175 & \cellcolor[RGB]{255,200,200}0.5287 & \cellcolor[RGB]{255,156,156}0.6545 & \cellcolor[RGB]{255,123,123}0.6379 & \cellcolor[RGB]{255,28,28}0.9547 & \cellcolor[RGB]{255,35,35}0.7882 \\
\hline
NLinear & \cellcolor[RGB]{165,223,166}0.2787 & \cellcolor[RGB]{112,203,113}0.3401 & \cellcolor[RGB]{255,168,168}0.5222 & \cellcolor[RGB]{255,193,193}0.5322 & \cellcolor[RGB]{255,145,145}0.6647 & \cellcolor[RGB]{255,114,114}0.6424 & \cellcolor[RGB]{255,4,4}0.9844 & \cellcolor[RGB]{255,14,14}0.8027 \\
\hline
Crossformer & \cellcolor[RGB]{32,175,35}\textbf{0.1633} & \cellcolor[RGB]{61,185,63}0.3042 & \cellcolor[RGB]{120,207,122}0.3203 & \cellcolor[RGB]{88,195,90}0.4142 & \cellcolor[RGB]{150,217,151}0.4432 & \cellcolor[RGB]{110,203,112}0.4919 & \cellcolor[RGB]{119,206,120}0.4746 & \cellcolor[RGB]{145,216,147}0.5486 \\
\hline
TiDE & \cellcolor[RGB]{164,222,165}0.2774 & \cellcolor[RGB]{110,203,112}0.3393 & \cellcolor[RGB]{255,171,171}0.5199 & \cellcolor[RGB]{255,193,193}0.5322 & \cellcolor[RGB]{255,138,138}0.6717 & \cellcolor[RGB]{255,104,104}0.6473 & \cellcolor[RGB]{255,11,11}0.9754 & \cellcolor[RGB]{255,17,17}0.8010 \\
\hline
SparseTSF & \cellcolor[RGB]{161,221,163}0.2755 & \cellcolor[RGB]{104,201,106}0.3345 & \cellcolor[RGB]{255,160,160}0.5287 & \cellcolor[RGB]{255,189,189}0.5339 & \cellcolor[RGB]{255,135,135}0.6751 & \cellcolor[RGB]{255,110,110}0.6441 & \cellcolor[RGB]{255,6,6}0.9822 & \cellcolor[RGB]{255,16,16}0.8016 \\
\hline
PatchTST & \cellcolor[RGB]{158,220,159}0.2725 & \cellcolor[RGB]{111,203,113}0.3396 & \cellcolor[RGB]{255,184,184}0.5089 & \cellcolor[RGB]{255,200,200}0.5287 & \cellcolor[RGB]{255,135,135}0.6744 & \cellcolor[RGB]{255,101,101}0.6488 & \cellcolor[RGB]{255,9,9}0.9779 & \cellcolor[RGB]{255,20,20}0.7987 \\
\hline
Autoformer & \cellcolor[RGB]{242,250,242}0.3452 & \cellcolor[RGB]{202,236,203}0.4043 & \cellcolor[RGB]{255,0,0}0.6647 & \cellcolor[RGB]{255,20,20}0.6137 & \cellcolor[RGB]{255,0,0}0.8061 & \cellcolor[RGB]{255,0,0}0.6987 & \cellcolor[RGB]{255,224,224}0.7075 & \cellcolor[RGB]{255,209,209}0.6670 \\
\hline
TimeMixer & \cellcolor[RGB]{130,210,132}0.2481 & \cellcolor[RGB]{92,196,94}0.3262 & \cellcolor[RGB]{111,203,113}0.3114 & \cellcolor[RGB]{131,211,133}0.4374 & \cellcolor[RGB]{218,242,219}0.5193 & \cellcolor[RGB]{255,234,234}0.5829 & \cellcolor[RGB]{255,143,143}0.8100 & \cellcolor[RGB]{255,131,131}0.7211 \\
\hline
WaveNet & \cellcolor[RGB]{255,0,0}0.5504 & \cellcolor[RGB]{255,0,0}0.6005 & \cellcolor[RGB]{31,175,34}\textbf{\textcolor{white}{0.2340}} & \cellcolor[RGB]{31,175,34}\textbf{\textcolor{white}{0.3834}} & \cellcolor[RGB]{143,215,145}0.4366 & \cellcolor[RGB]{137,213,139}0.5070 & \cellcolor[RGB]{150,217,151}0.5195 & \cellcolor[RGB]{152,218,154}0.5541 \\
\hline
FEDformer & \cellcolor[RGB]{110,203,112}0.2312 & \cellcolor[RGB]{130,210,131}0.3529 & \cellcolor[RGB]{134,211,135}0.3334 & \cellcolor[RGB]{169,224,170}0.4575 & \cellcolor[RGB]{130,210,131}0.4215 & \cellcolor[RGB]{216,241,217}0.5515 & \cellcolor[RGB]{255,151,151}0.7998 & \cellcolor[RGB]{255,70,70}0.7635 \\
\hline
iTransformer & \cellcolor[RGB]{209,238,209}0.3166 & \cellcolor[RGB]{181,228,182}0.3895 & \cellcolor[RGB]{255,2,2}0.6624 & \cellcolor[RGB]{255,0,0}0.6231 & \cellcolor[RGB]{255,101,101}0.7076 & \cellcolor[RGB]{255,82,82}0.6579 & \cellcolor[RGB]{255,0,0}0.9902 & \cellcolor[RGB]{255,0,0}0.8129 \\
\hline
TimeFlex & \cellcolor[RGB]{65,187,68}0.1919 & \cellcolor[RGB]{57,184,60}0.3017 & \cellcolor[RGB]{107,202,109}0.3076 & \cellcolor[RGB]{118,206,120}0.4304 & \cellcolor[RGB]{87,195,89}0.3738 & \cellcolor[RGB]{116,205,117}0.4949 & \cellcolor[RGB]{31,175,34}\textbf{\textcolor{white}{0.3485}} & \cellcolor[RGB]{31,175,34}\textbf{\textcolor{white}{0.4575}} \\
\hline
\end{tabular}
}
\caption{Full results (MSE and MAE) of different models trained on the ETTm2 dataset.} \label{tab:ETTm2}
\end{table}

%% file: main.bbl
\begin{thebibliography}{10}
\providecommand{\url}[1]{#1}
\csname url@samestyle\endcsname
\providecommand{\newblock}{\relax}
\providecommand{\bibinfo}[2]{#2}
\providecommand{\BIBentrySTDinterwordspacing}{\spaceskip=0pt\relax}
\providecommand{\BIBentryALTinterwordstretchfactor}{4}
\providecommand{\BIBentryALTinterwordspacing}{\spaceskip=\fontdimen2\font plus
\BIBentryALTinterwordstretchfactor\fontdimen3\font minus
  \fontdimen4\font\relax}
\providecommand{\BIBforeignlanguage}[2]{{%
\expandafter\ifx\csname l@#1\endcsname\relax
\typeout{** WARNING: IEEEtran.bst: No hyphenation pattern has been}%
\typeout{** loaded for the language `#1'. Using the pattern for}%
\typeout{** the default language instead.}%
\else
\language=\csname l@#1\endcsname
\fi
#2}}
\providecommand{\BIBdecl}{\relax}
\BIBdecl

\bibitem{DLinear}
A.~Zeng, M.~Chen, L.~Zhang, and Q.~Xu, ``Are transformers effective for time
  series forecasting?'' \emph{AAAI}, vol.~37, no.~9, pp. 11\,121--11\,128, Jun.
  2023.

\bibitem{RLinear}
\BIBentryALTinterwordspacing
Z.~Li, S.~Qi, Y.~Li, and Z.~Xu, ``Revisiting long-term time series forecasting:
  An investigation on linear mapping,'' 2023. [Online]. Available:
  \url{https://arxiv.org/abs/2305.10721}
\BIBentrySTDinterwordspacing

\bibitem{waveaug2024}
\BIBentryALTinterwordspacing
D.~Arabi, J.~Bakhshaliyev, A.~Coskuner, K.~Madhusudhanan, and K.~S. Uckardes,
  ``Wave-mask/mix: Exploring wavelet-based augmentations for time series
  forecasting,'' 2024. [Online]. Available:
  \url{https://arxiv.org/abs/2408.10951}
\BIBentrySTDinterwordspacing

\bibitem{TimeMixer}
S.~Wang, H.~Wu, X.~Shi, T.~Hu, H.~Luo, L.~Ma, J.~Y. Zhang, and J.~Zhou,
  ``Timemixer: Decomposable multiscale mixing for time series forecasting,''
  \emph{CoRR}, vol. abs/2405.14616, 2024.

\bibitem{TimesNet}
H.~Wu, T.~Hu, Y.~Liu, H.~Zhou, J.~Wang, and M.~Long, ``Timesnet: Temporal
  2d-variation modeling for general time series analysis,'' in \emph{ICLR},
  2023.

\bibitem{LSTM}
S.~Hochreiter and J.~Schmidhuber, ``{Long Short-Term Memory},'' \emph{Neural
  Computation}, vol.~9, no.~8, pp. 1735--1780, 1997.

\bibitem{GRU}
K.~Cho, B.~van Merri{\"e}nboer, C.~Gulcehre, D.~Bahdanau, F.~Bougares,
  H.~Schwenk, and Y.~Bengio, ``Learning phrase representations using {RNN}
  encoder{--}decoder for statistical machine translation,'' in
  \emph{Proceedings of the 2014 Conference on Empirical Methods in Natural
  Language Processing ({EMNLP})}, 2014, pp. 1724--1734.

\bibitem{CNN}
Y.~Zheng, Q.~Liu, E.~Chen, Y.~Ge, and J.~L. Zhao, ``Time series classification
  using multi-channels deep convolutional neural networks,'' in \emph{Web-Age
  Information Management}.\hskip 1em plus 0.5em minus 0.4em\relax Springer
  International Publishing, 2014, pp. 298--310.

\bibitem{WaveNet}
\BIBentryALTinterwordspacing
A.~van~den Oord, S.~Dieleman, H.~Zen, K.~Simonyan, O.~Vinyals, A.~Graves,
  N.~Kalchbrenner, A.~Senior, and K.~Kavukcuoglu, ``Wavenet: A generative model
  for raw audio,'' 2016. [Online]. Available:
  \url{https://arxiv.org/abs/1609.03499}
\BIBentrySTDinterwordspacing

\bibitem{TCN}
C.~Lea, M.~D. Flynn, R.~Vidal, A.~Reiter, and G.~D. Hager, ``Temporal
  convolutional networks for action segmentation and detection,'' in
  \emph{CVPR}, 2017, pp. 1003--1012.

\bibitem{Adawavenet}
H.~Yu, P.~Guo, and A.~Sano, ``Adawavenet: Adaptive wavelet network for time
  series analysis,'' \emph{Transactions on Machine Learning Research}, 2024.

\bibitem{Li_transformer}
S.~Li, X.~Jin, Y.~Xuan, X.~Zhou, W.~Chen, Y.-X. Wang, and X.~Yan, ``Enhancing
  the locality and breaking the memory bottleneck of transformer on time series
  forecasting,'' in \emph{NeurIPS}, vol.~32, 2019.

\bibitem{Informer}
H.~Zhou, S.~Zhang, J.~Peng, S.~Zhang, J.~Li, H.~Xiong, and W.~Zhang,
  ``Informer: Beyond efficient transformer for long sequence time-series
  forecasting,'' \emph{AAAI}, vol.~35, no.~12, pp. 11\,106--11\,115, May 2021.

\bibitem{Autoformer}
H.~Wu, J.~Xu, J.~Wang, and M.~Long, ``Autoformer: decomposition transformers
  with auto-correlation for long-term series forecasting,'' in \emph{NeurIPS},
  2021.

\bibitem{FEDFormer}
T.~Zhou, Z.~Ma, Q.~Wen, X.~Wang, L.~Sun, and R.~Jin, ``{FED}former: Frequency
  enhanced decomposed transformer for long-term series forecasting,'' in
  \emph{ICML}, 2022, pp. 27\,268--27\,286.

\bibitem{Pyraformer}
S.~Liu, H.~Yu, C.~Liao, J.~Li, W.~Lin, A.~X. Liu, and S.~Dustdar, ``Pyraformer:
  Low-complexity pyramidal attention for long-range time series modeling and
  forecasting,'' in \emph{ICLR}, 2022.

\bibitem{Crossformer}
Y.~Zhang and J.~Yan, ``Crossformer: Transformer utilizing cross-dimension
  dependency for multivariate time series forecasting,'' in \emph{ICLR}, 2023.

\bibitem{PatchTST}
Y.~Nie, N.~H. Nguyen, P.~Sinthong, and J.~Kalagnanam, ``A time series is worth
  64 words: Long-term forecasting with transformers,'' in \emph{ICLR}, 2023.

\bibitem{Itransformer}
Y.~Liu, T.~Hu, H.~Zhang, H.~Wu, S.~Wang, L.~Ma, and M.~Long, ``itransformer:
  Inverted transformers are effective for time series forecasting,'' in
  \emph{ICLR}, 2024.

\bibitem{VCFormer}
Y.~Yang, Q.~Zhu, and J.~Chen, ``Vcformer: Variable correlation transformer with
  inherent lagged correlation for multivariate time series forecasting,'' in
  \emph{IJCAI}, 2024, pp. 5335--5343.

\bibitem{wang2024timexer}
Y.~Wang, H.~Wu, J.~Dong, Y.~Liu, Y.~Qiu, H.~Zhang, J.~Wang, and M.~Long,
  ``Timexer: Empowering transformers for time series forecasting with exogenous
  variables,'' \emph{NeurIPS}, 2024.

\bibitem{Mamba}
\BIBentryALTinterwordspacing
A.~Gu and T.~Dao, ``Mamba: Linear-time sequence modeling with selective state
  spaces,'' 2024. [Online]. Available: \url{https://arxiv.org/abs/2312.00752}
\BIBentrySTDinterwordspacing

\bibitem{TimeMachine}
M.~A. Ahamed and Q.~Cheng, ``Timemachine: A time series is worth 4 mambas for
  long-term forecasting,'' in \emph{ECAI}, vol. 392, 2024, pp. 1688--1695.

\bibitem{TiDE}
A.~Das, W.~Kong, A.~Leach, S.~K. Mathur, R.~Sen, and R.~Yu, ``Long-term
  forecasting with ti{DE}: Time-series dense encoder,'' \emph{Transactions on
  Machine Learning Research}, 2023.

\bibitem{Sparsetsf}
S.~Lin, W.~Lin, W.~Wu, H.~Chen, and J.~Yang, ``Sparsetsf: Modeling long-term
  time series forecasting with 1k parameters,'' in \emph{ICML}, 2024.

\bibitem{Cyclenet}
S.~Lin, W.~Lin, X.~Hu, W.~Wu, R.~Mo, and H.~Zhong, ``Cyclenet: Enhancing time
  series forecasting through modeling periodic patterns,'' in \emph{NeurIPS},
  2024.

\bibitem{MoU}
\BIBentryALTinterwordspacing
S.~Peng, Y.~Xiong, Y.~Zhu, and Z.~Shen, ``Mamba or transformer for time series
  forecasting? mixture of universals (mou) is all you need,'' 2024. [Online].
  Available: \url{https://arxiv.org/abs/2408.15997}
\BIBentrySTDinterwordspacing

\bibitem{Onenet}
Y.~Zhang, Q.~Wen, X.~Wang, W.~Chen, L.~Sun, Z.~Zhang, L.~Wang, R.~Jin, and
  T.~Tan, ``Onenet: Enhancing time series forecasting models under concept
  drift by online ensembling,'' in \emph{NeurIPS}, 2023.

\bibitem{wang2024survey}
\BIBentryALTinterwordspacing
Y.~Wang, H.~Wu, J.~Dong, Y.~Liu, M.~Long, and J.~Wang, ``Deep time series
  models: A comprehensive survey and benchmark,'' 2024. [Online]. Available:
  \url{https://arxiv.org/abs/2407.13278}
\BIBentrySTDinterwordspacing

\bibitem{qiu2024tfb}
X.~Qiu, J.~Hu, L.~Zhou, X.~Wu, J.~Du, B.~Zhang, C.~Guo, A.~Zhou, C.~S. Jensen,
  Z.~Sheng, and B.~Yang, ``Tfb: Towards comprehensive and fair benchmarking of
  time series forecasting methods,'' \emph{Proc. {VLDB} Endow.}, vol.~17,
  no.~9, pp. 2363--2377, 2024.

\bibitem{RasmussenWilliams2005}
C.~E. Rasmussen and C.~K.~I. Williams, \emph{Gaussian Processes for Machine
  Learning}, ser. Adaptive Computation and Machine Learning Series.\hskip 1em
  plus 0.5em minus 0.4em\relax Cambridge, MA: The MIT Press, 2005.

\bibitem{kim2022reversible}
T.~Kim, J.~Kim, Y.~Tae, C.~Park, J.-H. Choi, and J.~Choo, ``Reversible instance
  normalization for accurate time-series forecasting against distribution
  shift,'' in \emph{ICLR}, 2022.

\bibitem{PixelCNN}
A.~van~den Oord, N.~Kalchbrenner, L.~Espeholt, K.~Kavukcuoglu, O.~Vinyals, and
  A.~Graves, ``Conditional image generation with pixelcnn decoders,'' in
  \emph{NeurIPS}, vol.~29, 2016.

\end{thebibliography}
